\documentclass[preprint,12pt]{elsarticle}

\usepackage{amssymb}
\usepackage{amsmath}
\usepackage[ruled,vlined,linesnumbered]{algorithm2e}            
\biboptions{square,numbers,sort&compress}
\usepackage{booktabs}
\usepackage{makecell}
\usepackage{multicol}
\usepackage{multirow}
\usepackage{tabularx}
\usepackage{caption}
\usepackage{float}
\usepackage{graphicx}
\usepackage{longtable}
\usepackage{xltabular}
\usepackage{enumitem}
\usepackage{tabularx}
\usepackage{geometry}
\usepackage{rotating}
\usepackage{threeparttable}
\usepackage{array} 
\usepackage{afterpage}

\newcolumntype{C}[1]{>{\centering\arraybackslash}p{#1}}
\usepackage{booktabs} 
\newcolumntype{R}[1]{>{\raggedright\arraybackslash}p{#1}}
\usepackage{caption} 
\graphicspath{ {./Figs/} }

\journal{Knowledge-Based Systems}

\begin{document}

\begin{frontmatter}



\title{ARM-Explainer - Explaining and improving graph neural network predictions for the maximum clique problem using node features and association rule mining}


\author[inst1]{Bharat S. Sharman\fnmark[1]}
\affiliation[inst1]{organization={School of Computational Science and Engineering},
            addressline={McMaster University}, 
            city={Hamilton},
            postcode={L8S 4E8}, 
            state={ON},
            country={Canada}}

\author[inst2]{Elkafi Hassini}
\affiliation[inst2]{organization={DeGroote School of Business},
            addressline={McMaster University}, 
            city={Hamilton},
            postcode={L8S 4E8}, 
            state={ON},
            country={Canada}}

\fntext[1]{Corresponding author: sharmanb@mcmaster.ca (Bharat S. Sharman)}

\begin{abstract}
Numerous GNN-based algorithms have been proposed to solve graph-based combinatorial optimization problems (COPs). Yet, methods to explain the predictions of these algorithms for graph-based COPs have not yet been developed. This research introduces ARM-Explainer, a post-hoc, model-level explainer utilizing association rule mining, demonstrated on the hybrid geometric scattering (HGS) GNN algorithm's predictions for the maximum clique problem (MCP), a recognized NP-hard graph-based COP. The eight top explainable association rules generated by the ARM-Explainer had high median lift and confidence values of 2.42 and 0.49, respectively, on test instances from the TWITTER and BHOSLIB-DIMACS benchmark datasets. It identified the important node features and their ranges that influenced the GNN's predictions for these datasets. Moreover, incorporating informative node features into the GNN notably enhanced its predictive capability for the MCP, increasing the median largest-found clique size by 22\% (from 29.5 to 36) for large graphs of the BHOSLIB-DIMACS dataset.    
\end{abstract}

\begin{graphicalabstract}
\includegraphics[width=15.5cm,height=12cm]{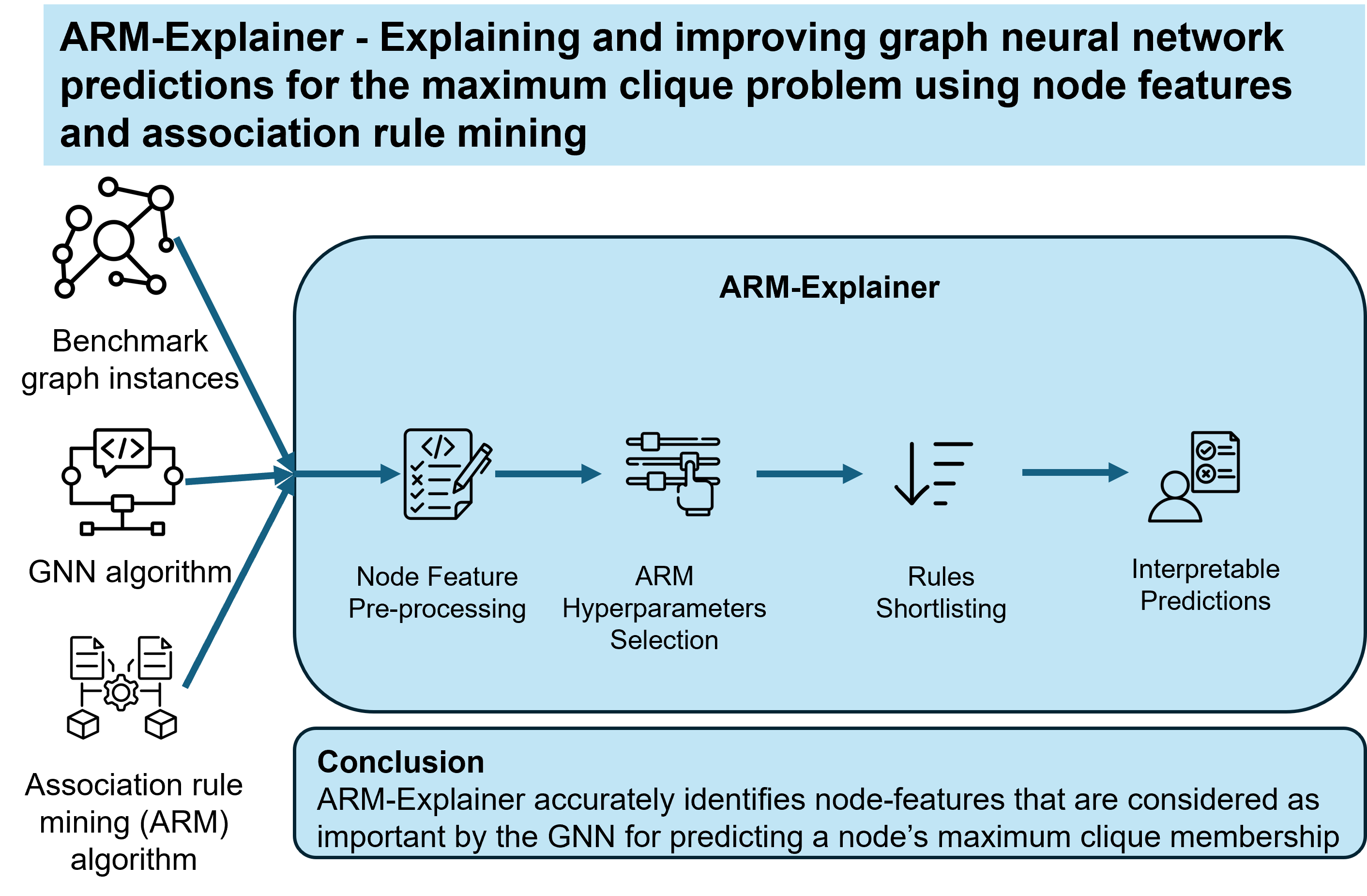}
\end{graphicalabstract}


\begin{highlights}
\item We introduce ARM-Explainer for explainable ML in combinatorial optimization.
\item Explains predictions of a hybrid scattering GNN on the maximum clique problem.
\item Adding informative node features improves GNN accuracy, especially on large graphs.
\item ARM-Explainer finds key node features across inputs and graph sizes.
\end{highlights}


\end{frontmatter}



\section{Introduction}
\label{sec:intro}

Combinatorial optimization problems (COPs) involve selecting a subset of objects from a finite set such that the selected subset optimizes an objective function defined over that set, subject to specific constraints \cite{du2022introduction}. COPs are of immense theoretical and practical importance. For example, it is well known that several COPs, such as the traveling salesperson problem (TSP), are $\cal{NP}$-hard and are widely used to formulate and solve problems in several areas, such as supply chain management \cite{hassini2025modeling}, logistics \cite{mansouri2025freight}, bioinformatics \cite{kaya2022review}, chemical engineering \cite{mizuno2024finding} and cybersecurity \cite{dragotto2024critical},  to name but a few. \\
A number of combinatorial optimization problems (COPs) such as the maximum clique (MCP), maximum independent set (MIS), minimum vertex cover (MVC), maximum cut (MaxCut), graph coloring (GCP), vehicle routing (VRP), and location routing (LRP) can be modeled using graphs and are collectively known as graph-based COPs. Many of these problems are $\cal{NP}$-hard, exhibiting exponential complexity in the worst-case scenario as the problem size grows. This paper focuses on the Maximum Clique Problem (MCP), which seeks to determine the largest fully connected subgraph within a graph \cite{wu2015review}. Karp's seminal work \cite{karp1972} included the decision variant of the MCP among the initial 21 $\cal{NP}$-complete problems. The MCP is of significant theoretical and practical interest, prompting extensive research. Various methods have been proposed for solving the MCP, including exact algorithms with branch and bound (B\&B) techniques \citep{li2017minimization,li2010efficient,jiang2016combining,san2019new,san2023clisat}, greedy and local search techniques \citep{wu2012multi,wang2016two,wang2020sccwalk}, Monte Carlo simulations \cite{angelini2021mismatching}, and, more recently, graph neural networks (GNN) \citep{karalias2020erdos,min2022can,sanokowski2024variational}.
Approaches such as branch and bound, greedy and local search, and Monte Carlo methods are characterized by the transparency of their algorithmic steps. On the other hand, explaining the results of GNNs is challenging due to the complex transformations performed by multiple layers of neural networks on the inputs to generate the outputs. While several approaches have been proposed in the literature to explain the predictions of GNNs for a variety of node and graph classification tasks, work on the explainability of these methods for combinatorial problems has been absent \cite{kakkad2023survey}. To the best of the authors' knowledge, there has been no prior work systematically explaining the predictions of the GNNs for the MCP or for graph-based COPs in general. Additionally, existing studies on GNN methods for solving graph-based COPs typically consider very few (mostly up to 3) node features in their analysis \citep{karalias2020erdos,min2022can,wenkel2024towards}, and the effect of adding more node features on the performance of GNNs on graph-based COPs has not been investigated. 

The main contributions of this study are, therefore, the following:

\begin{itemize}

    \item Methodology for explaining the predictions of GNNs for graph-based COPs: We propose a methodology based on association rule mining (ARM) that we term ARM-Explainer to explain the predictions of GNNs for graph-based COPs. The generic nature of this methodology makes it suitable for application to a variety of GNN algorithms and graph-based COPs.   
    \item Investigating the effect of additional node features on GNNs predictive performance for graph-based COPs : We investigate whether and to what extent the predictive performance of GNNs changes when additional node features are provided as inputs. 
    \item Extending GNN predictions to benchmark larger graph datasets : GNN algorithms have typically been applied on smaller-sized graph benchmarks such as IMDB-BINARY and COLLAB. We extend them to larger graph benchmarks such as BHOSLIB and DIMACS.  

\end{itemize}

This work seeks to offer insights into enhancing the explainability and performance of GNN predictions for graph-based COPs, targeting both academic and industrial users. The paper is structured as follows: Section \ref{sec:litreview} reviews methods for explaining GNNs. Section \ref{sec:mcp} discusses the maximum clique problem and outlines the hybrid geometric scattering (HGS) GNN algorithm used to demonstrate our method. Section \ref{sec:fpg_explanation_for_mcp} details the ARM-Explainer, along with datasets, node features, and computational aspects. Results are shown in section \ref{sec:results}, while Section \ref{sec:conclusion} summarizes key findings and suggests future research paths.    

\subsection{Graph Neural Networks (GNNs)}
\label{subsec:gnn_intro}

GNNs represent a class of neural networks adept at learning from graph-structured data. Consider an undirected, unweighted graph $G = (V, E)$, where $V$ and $E$ denote the set of $n$ nodes and $m$ edges respectively. The adjacency matrix \textbf{A} $\in$ $[0,1]^{n\times n}$ for $G$ is defined such that $A_{ij}=1$ if $(i,j) \in E$, otherwise 0. The nodes have features represented by the matrix \textbf{X} $\in$ $\mathbb{R}^{n\times F}$, where each row $i$ of \textbf{X} contains the $F$-dimensional feature vector for node $i$. A GNN model $\mathbb{M}$ maps each node into a reduced-dimensional space $\mathbb{Z} = \mathbb{R}^{n\times d}$ ($d < F$) through $l$-step message-passing given by:

\begin{equation}
\label{eq:gnn_formulation}
\begin{aligned}
    \mathbb{Z}_{v}^{l+1} = UPDATE_{\Phi}(\mathbb{Z}_{v}^{l},AGG_{\Psi}((MSG_{\Theta}(\mathbb{Z}_{v}^{l},\mathbb{Z}_{u}^{l} : (u,v) \in E)) \\
\end{aligned}
\end{equation}

starting with $\mathbb{Z}^{0} = \textbf{X}$ and resulting in $\mathbb{Z}$ := $\mathbb{Z}^{l}$. Various GNN architectures, including graph convolutional networks (GCN) \cite{kipf2016semi}, GraphSAGE \cite{hamilton2017inductive}, and graph attention networks (GAT) \cite{velivckovic2017graph}, arise from distinct combinations of update ($UPDATE_{\Phi}$), aggregation ($AGG_{\Psi}$), and message generation ($MSG_{\Theta}$) functions. The weights for these functions across GNN layers can be trained for diverse tasks, such as node and graph classification, link prediction, clustering, and temporal dynamics prediction.

\section{Literature review}
\label{sec:litreview}

In recent years, GNN research has experienced significant expansion due to their extensive use across diverse domains, such as bioinformatics \citep{pfeifer2022gnn,metsch2024clarus,mastropietro2023xgdag}, chemical and molecular property prediction \citep{yang2022mgraphdta,tian2023predicting,aouichaoui2023application,jian2022predicting,kotobi2023integrating,low2022explainable,xie2018crystal,yang2022learning,henderson2021improving,proietti2024explainable,liu2023interpretable,li2023predicting,rao2022quantitative,aouichaoui2023combining,dai2021graph}, computational fluid dynamics \citep{barwey2023multiscale}, cybersecurity \citep{ganz2021explaining,he2022illuminati,zhu2022interpretability,lo2023xg,warmsley2022survey,cao2024coca}, deepfake detection \citep{khalid2023dfgnn}, e-commerce \citep{amara2022graphframex}, energy \citep{gao2022interpretable,verdone2024explainable}, forecasting \citep{li2022online}, geolocation \citep{zhou2022identifying}, medicine \citep{mokhtari2022echognn,gao2024aligning,sihag2023explainable}, natural language processing \citep{christmann2023explainable}, neuroscience \citep{li2021braingnn,cui2021brainnnexplainer,cui2022interpretable,zheng2024ci,li2023interpretable}, process monitoring \citep{harl2020explainable}, recommender systems \citep{lyu2022knowledge,shuai2023topic}, robotics \citep{lin2022efficient}, climate and weather prediction \citep{yang2024interpretable}, and transportation \citep{tygesen2023unboxing}. The critical impact of predictions in domains such as healthcare, drug discovery, and cybersecurity has driven the GNN research community to create several methods to explain GNN predictions.\\

A comprehensive survey of GNN explainability methods has been provided in  \cite{kakkad2023survey}. The literature survey presented here is based on this work and extends it to include new approaches that have since been proposed. Methods for explaining the predictions of GNNs can be broadly classified as factual and counterfactual. Factual algorithms aim to directly find the input features of the graph instance that have the maximum influence on the prediction of the GNN. These features can be either
nodes, edges, subgraphs, higher-order structures such as cells, or a combination of these. On the other
hand, counterfactual algorithms alter various features and compute the effects that these changes have on the predictions of the GNN. The features whose alteration leads to significant changes in predictions are identified as explanatory features. The explanations can be instance-level, where the explainer provides explanations for a particular graph instance, or they can be global or model-level; in which case, the explainer extracts
patterns utilized by the GNN across several graphs to generate explanations. \cite{armgaan2024graphtrail} \\
Factual algorithms can be further categorized into post-hoc and self-interpretable algorithms. 
Post-hoc algorithms use a model that is separate from the GNN, which takes the inputs fed to the GNN, its output, and in some cases, its internal parameters, and then outputs the features that it determines are the explanatory factors behind the predictions. Different models use different parts of the inputs utilized by the GNN to determine the explanatory factors. Post-hoc algorithms can be further classified into decomposition \citep{pope2019explainability,feng2023degree,schnake2021higher,lu2024goat}, gradient \citep{baldassarre2019explainability,lu2024eig,pope2019explainability}, surrogate \citep{akkas2024gnnshap,duval2021graphsvx,huang2022graphlime,luo2020parameterized,pereira2023distill,zhang2021relex,chang2024path,armgaan2024graphtrail,azzolin2023global,zhao2025multi}, perturbation \citep{luo2020parameterized,bui2024explaining,chen2024generating,funke2022zorro,schlichtkrull2020interpreting,wang2021towards,ying2019gnnexplainer,yuan2021explainability,zhang2022gstarx}, and generation \citep{yuan2020xgnn,wang2022gnninterpreter,shan2021reinforcement,lin2021generative,li2023dag,wang2024gnnboundary,gaines2025explaining,saha2024graphon,hu2025dgx} based methods. 

Self-interpretable algorithms, on the other hand, do not have a separate module for explanation; instead, it is built into the GNN itself. When a graph instance is fed as input to the GNN for prediction, the explanation module extracts a sub-graph from the input graph and then uses this sub-graph for both prediction and explanation. Therefore, self-interpretable algorithms have a trade-off between higher explainability and lower prediction accuracy. Self-interpretable algorithms can be further classified into those that are based on information constraints \citep{huang2024factorized,ji2024stratified,miao2022interpretable,muller2024graphchef,yu2020graph,yu2022improving} and those that are based on structural constraints \citep{zhang2022protgnn,wu2022discovering,feng2022kergnns,deng2024self,dai2021towards}.\\

Counterfactual algorithms can be further classified into search-based, neural network-based, and perturbation-based algorithms, respectively. Search-based algorithms \citep{chen2024view,chhablani2024game,numeroso2021meg,wellawatte2022model,kosan2025gcfexplainer} search the counterfactual space for examples that can alter the predictions of the GNN. They are useful in situations where the prediction for an input is definitively known, and it is required to alter the input so that one obtains a different prediction. The challenge that search-based counterfactual algorithms face is that they need to be efficient, as the counterfactual search space can be exponential in size. Neural network-based algorithms \citep{ma2022clear,bajaj2021robust} use a neural network to generate a counterfactual instance from an existing instance. Neural networks can be used for either generating edges between nodes or generating a complete graph. Perturbation-based methods \citep{chen2024feature,chen2024interpretable,chen2022grease,kangunr2024,lucic2022cf,qiu2024generating,tan2022learning,vermainduce2024} add or delete edges to change the prediction of the GNN and do so either by altering the adjacency matrix or the computational graph of a node.\\

Based on the aforementioned review,this work falls into the category of factual, post-hoc methods and uses Frequent Pattern Growth (FP-Growth), an ARM method, as the surrogate model to generate model-level explanations for GNNs predictions of graph-based COPs.

\section{The maximum clique problem}
\label{sec:mcp}

\subsection{Definitions and notations}
\label{subsec:defn_and_notation}

Let $G = (V,E)$ be an undirected and unweighted graph, where \(V \) is the set of nodes and \(E \) is the set of edges of the graph. The vertex cardinality of a graph is the number of vertices and is indicated by $|V|$, while the edge cardinality of a graph is the number of edges and is denoted by $|E|$. For a subset of nodes $S \subseteq V$, the set $G(S) = G(S,E \subseteq S \times S)$ is known as the sub-graph induced by S. A complete graph is a graph where every pair of vertices is adjacent, i.e., $\forall (i,j) \in V$; if $i \neq j$, then $(i,j) \in E $. A clique $C$ is a complete subgraph of $G$. A clique is called maximal if it is not a subset of a larger clique in $G$. A clique is said to be the maximum clique if its vertex cardinality is the highest among all the cliques in the graph. The MCP can be formulated as the following integer linear programming problem:

\begin{equation}
\label{eq:mcp_ilp_formulation}
\begin{aligned}
    \text{Maximize } & \sum_{i \in V} x_i \\
    \text{subject to: } & x_i + x_j \leq 1 \quad \forall (i, j) \notin E \\
    & x_i \in \{0, 1\} \quad \forall i \in V
\end{aligned}
\end{equation}

where \( x_i \) is a binary variable that indicates whether the vertex \( i \) is included in the clique (\( x_i = 1 \)) or not (\( x_i = 0 \)). The inequality constraint in Equation \ref{eq:mcp_ilp_formulation} specifies the condition that if there is no edge between two nodes in the graph, they cannot both be members of a clique. The clique number $\omega (G)$ is the number of vertices in the maximum clique of G. In addition to this classic version of the MCP defined on an unweighted graph, weighted versions of the MCP also exist. If each node $i$ of a graph has a weight $w_{i}$ associated with it, these weights can collectively be represented by a weight vector $ \vec w \in R^{N} $. The Maximum Weighted Clique Problem (MWCP) requires finding the clique with the maximum sum of node weights. In an analogous manner, the maximum weighted clique problem can also involve finding the clique with the maximum sum of edge weights instead of node weights. The focus of this work is on the classic version of the MCP.
        
\subsection{Hybrid geometric scattering (HGS) algorithm}
\label{subsec:hgs_algorithm}

The algorithm that we considered to demonstrate our GNN explainability method is the HGS GNN proposed by \cite{min2022can}. HGS is a self-supervised algorithm that trains a geometric deep learning network based on a customized loss function to maximize its performance on the MCP. It was shown to achieve state of the art performance compared to EGN \cite{karalias2020erdos}, another GNN-based algorithm, as well as mixed integer programming (MIP)-based Gurobi on graph-based COP benchmark datasets such as TWITTER \cite{yan2008mining}, IMDB, and COLLAB\cite{yanardag2015deep}. In contrast to exact algorithms such as Gurobi \cite{gurobi}, CliSAT \cite{san2023clisat}, or MoMC \cite{li2017minimization}, which use a variant of the branch-and-bound technique to directly solve Equation \ref{eq:mcp_ilp_formulation} and output the largest found clique, GNN-based algorithms such as HGS take node features as inputs and output the probability of each node being a member of a maximum clique. The transformation of node features into node probabilities is accomplished through several layers of message passing and aggregation operations. They then use a rule-based decoder to output the largest found clique by selecting a subset of fully-connected nodes, starting with the node that has the highest computed probability. Such a highly non-linear transformation between inputs and outputs leads to the black-box nature of such models, which has, in turn, motivated several approaches to explain them, as discussed in Section \ref{sec:litreview}. We now provide a short introduction to the HGS algorithm. For further details, readers are referred to \cite{min2022can}. \\
\subsubsection{Graph scattering transform}
\label{subsec:graph_scattering_transform}

Let $G = (V,E)$ be an undirected and unweighted graph where $V := \{v_1,v_2,....v_n\}$ is the set of nodes and $E \subseteq V \times V$ is the set of edges of the graph. Let $\textbf{W} \in R^{n \times n}$ be the adjacency matrix and $\textbf{D} := diag\{d_1,d_2,....d_n\} \in R^{n \times n}$ be the degree matrix of $G$, where $d_{i} :=\sum_{j=1}^{n} W[v_{i},v_{j}]$ is the degree of node $i$. The lazy random walk matrix can be defined as:

\begin{equation}
\label{eq:lazy_random_walk_matrix}
\begin{aligned}
            \textbf{P}:=\frac{1}{2}(\textbf{I}_{n}+\textbf{W}\textbf{D}^{-1})
\end{aligned}
\end{equation}
where $\textbf{I}_{n}$ is the identity matrix. $\textbf{P}$ represents a diffusive process with self-retention; i.e., it models a random walker that has a nonzero probability of staying in place at a node. The graph scattering transform \citep{gama2018diffusion,zou2020graph,gao2019geometric} is based on the lazy random walk matrix and is a wavelet-based framework for learning on graphs. In contrast to GCNs, which apply one-hop localized low-pass filters to encourage feature smoothness among adjacent nodes, 
the graph scattering transform employs band-pass wavelet filters that capture long-range dependencies in a graph through their large spatial support. This is accomplished by raising the lazy random walk matrix defined in Equation \ref{eq:lazy_random_walk_matrix} to different powers, thereby encoding the diffusion geometry of the graph $G$ at multiple time scales. Subsequently, subtracting such consecutive power terms from each other makes it possible to highlight variations in diffusion geometry across scales. Following the approach of \cite{coifman2006diffusion}, 
for $k \in \mathbb{N}_0$, the wavelet matrix $\Psi_k \in \mathbb{R}^{n \times n}$ at scale $2^k$ is defined as
\begin{equation}
\label{eq:wavelet_matrix}
\Psi_0 = I_n - P, \qquad 
\Psi_k = P^{2^{k-1}} - P^{2^k}, \quad k \ge 1.
\end{equation}
Intuitively, for each node, these diffusion wavelets perform a comparison between averaged node features computed over two neighborhoods of different sizes, namely of size $2^{k-1}$ and $2^k$. This comparison reveals how node-level information evolves as it diffuses across increasingly larger neighborhoods in the graph.

\subsubsection{Components of HGS}
\label{subsec:hgs_components}
The HGS algorithm is composed of three key elements: a GNN-based model for deriving node probabilities, an unsupervised loss function tailored to refine the probability distribution over nodes for the MCP, and a rule-based decoder designed to identify the largest clique. The following is a brief overview of these components:
\begin{enumerate}
    \item 
\textbf{GNN-based training model}: The node features provided as input to the model are first embedded through an embedding module that transforms the features matrix $\textbf{X} \in R^{n \times d}$ into an embedding $\mathbf{H}^{0}$ of dimensions $d_{h}$ through a multi-layer perceptron (MLP) $m_{emb}$, $\mathbf{H}^{0}:=m_{emb}(\textbf{X})$. The authors who proposed the HGS model used three nodal features, namely eccentricity, the clustering coefficient, and the logarithm of the degree of each node as inputs. The embedding is then passed on to a diffusion module that has a series of $K \in \mathbb{N}$  diffusion layers. In each layer, the nodes have access to low-pass and band-pass filters, given by: 

\begin{equation}
\label{eq:low_and_bandpass_filters}
f_{low,r}(\mathbf{H}^{l-1}) = \mathbf{A}^{r}(\mathbf{H}^{l-1}), \qquad
f_{band,k}(\mathbf{H}^{l-1}) = \mathbf{\psi}_{k}(\mathbf{H}^{l-1})
\end{equation}

In Equation \ref{eq:low_and_bandpass_filters}, the low-pass filters apply the matrix $\textbf{A}:=(\textbf{D+I})^{-1/2}(\textbf{W+I})(\textbf{D+I})^{-1/2}$ raised to the power of $r \geq 1$, and the band-pass filters apply the wavelet matrix $\psi_{k}$ of order $k$, as defined in Equation \ref{eq:wavelet_matrix}, to the embeddings of the layer $l-1$ respectively. The importance of each of these filters to a node is computed through an attention mechanism based computation as follows:

\begin{equation}
\label{eq:filter_importance_scores}
s_{f}^{l} = \sigma ( \textbf{H}_{f}^{l} || \textbf{H}^{l-1} ) \textbf{a}^{l} 
\end{equation}
where $\textbf{H}_{f}^{l}:=f(\mathbf{H}^{l-1})$. The embeddings of the previous layer $l-1$ and the current layer $l$ are computed, concatenated, and a sigmoid transformation is applied to the result. Then a learned attention vector $\textbf{a}^{l}$ is applied to the result to compute the importance score of a filter for a node. The node representations are then recomputed as:
\begin{equation}
\label{eq:node_representation_update}
\textbf{H}_{agg}^{l} = \sum_{f \in F} \mathbf{\alpha}_{f}^{l} \odot \textbf{H}_{f}^{l}
\end{equation}
where $\odot$ is an element-wise multiplication operation and ${\alpha}_{f}(v)=softmax_{F}s_{f}(v)$. Such a softmax-based normalization ensures that the aggregate embedding $\textbf{H}_{agg}^{l}$ in Equation \ref{eq:node_representation_update} comprises contributions predominantly from one filter. The aggregated embedding is then passed through an MLP ($m^{l}$) to arrive at the final node representation for that layer, i.e., $\textbf{H}^{l} = m^{l} (\textbf{H}_{agg}^{l})$. The computed representations for all the $K$ layers are then concatenated to obtain $\textbf{H}_{cat}$, which is subsequently transformed into a vector $\textbf{h} \in \mathbb{R}^{n}$ through an MLP, i.e. $\textbf{h} = {m}_{out}(\textbf{H}_{cat})$. The probability vector for nodes to belong to a maximum clique is then obtained through min-max scaling applied to $\textbf{h}$,i.e., $\textbf{p} = (\textbf{h}-min(\textbf{h}) \cdot \textbf{1}_{n} )/(max(\textbf{h})-min(\textbf{h}))$.

\item \textbf{Custom supervised loss function:} To ensure that the set of nodes found is as large as possible and simultaneously satisfies the constraints of a maximum-clique, a custom loss function having two components has been proposed by the authors in \cite{min2022can}: 

\begin{equation}
\label{eq:custom_loss_function_for_mcp}
\textbf{L(p)} = \textbf{L}_{1} (\textbf{p}) + \mathbf(\beta) \textbf{L}_{2} (\textbf{p}) = -\textbf{p}^{T}\textbf{W}\textbf{p} + \mathbf(\beta)\textbf{p}^{T}\overline{ \textbf{W}}\textbf{p} 
\end{equation}

The intuition behind such a loss function is as follows: the first component $\textbf{L}_{1}$ assigns higher probabilities to those nodes that are highly connected. The second component has the adjacency matrix of the complement graph ($\overline{ \textbf{W}}$), which will have non-zero terms in case the nodes identified as part of a maximum clique are not connected to each other. In other words, it adds a penalty term whenever the maximum clique constraint is violated, and the relative importance of these loss terms can be adjusted by the hyperparameter $\beta$.     

\item \textbf{Rule-based decoder:} Once the model has been trained using the unsupervised loss function specified in Equation \ref{eq:custom_loss_function_for_mcp}, the output clique is finally obtained by applying a rule-based decoder. The detailed algorithmic steps of the decoder are provided \cite{min2022can}, and a brief overview of its steps is now presented. The decoder initially orders the nodes in decreasing order of probabilities obtained after the training step. The process of building the clique starts with the first node,i.e., the one that has the highest computed probability. The decoder then goes to the second node and checks whether it is connected to the first node. If yes, it adds it to the clique set. If not, it goes to the third node, and so on. At each node, it checks whether the node is connected to all the nodes in the existing clique set. It executes this loop for a defined number of times, as determined by a hyperparameter, and when all such cliques have been found, it outputs the clique with the maximum cardinality.     

\end{enumerate}
 
\section{ARM-Explainer - Explaining the predictions of HGS for the MCP using association rule mining}
\label{sec:fpg_explanation_for_mcp}

\begin{figure}[ht!]
    \centering
    \includegraphics[width=1.0\textwidth]{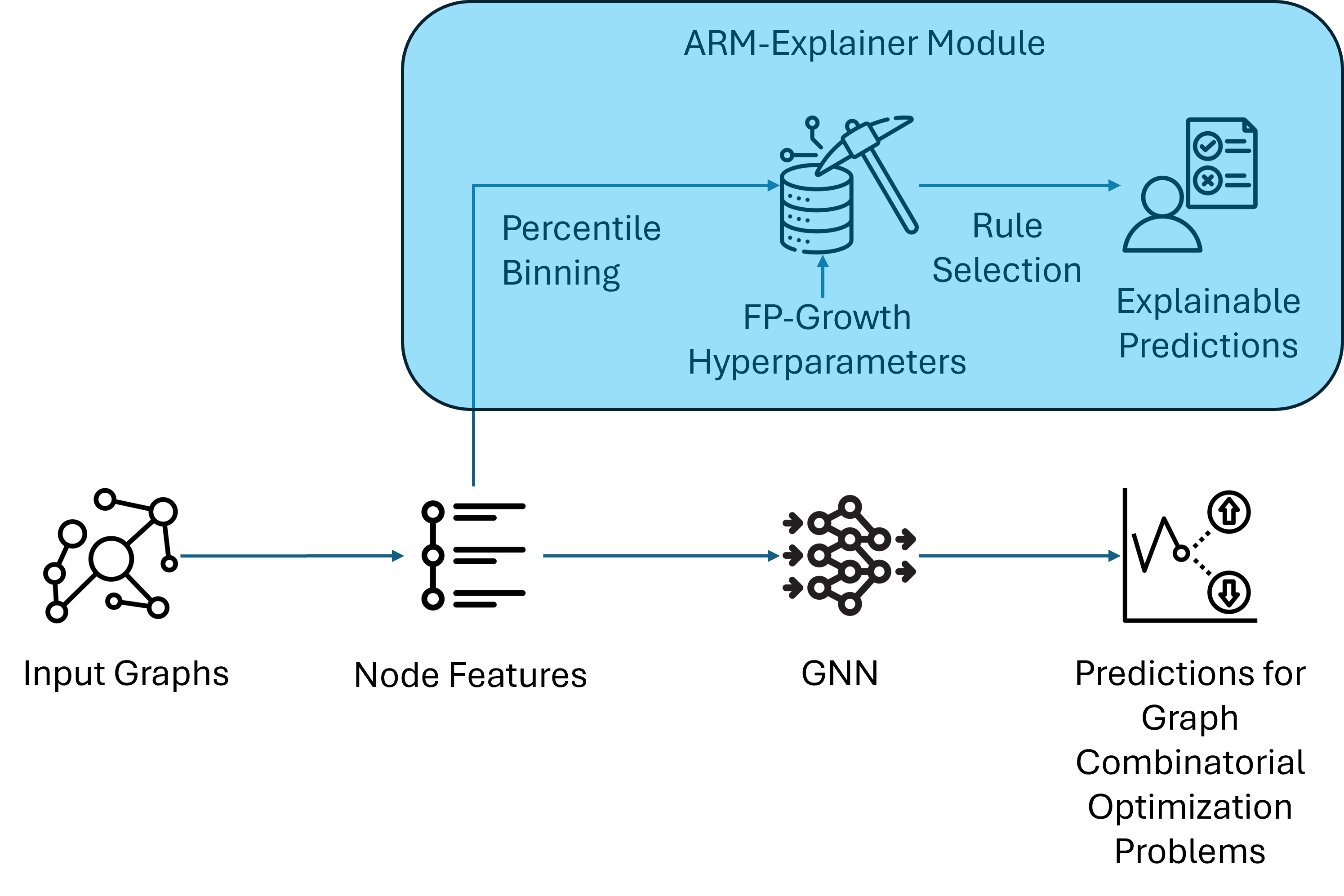}
    \caption{Illustration of the ARM-Explainer- The part highlighted in blue is the ARM-based explanation module. The node features and GNN predictions (node probabilities for belonging to a maximum clique) are used across all the graph instances are used to derive post-hoc model-level explainable rules that serve as explanations for GNN predictions}
    \label{fig:ARM_Explainer_Schematic}
\end{figure}

This section outlines ARM-Explainer, the methodology that we propose to explain the predictions of GNNs for graph-based COPs using association rule mining. Figure \ref{fig:ARM_Explainer_Schematic} shows the schematic of the ARM-Explainer. As a post-hoc model-level GNN explanation method, it takes in the node features of the input graphs and the predictions of a GNN for a graph-based COP, such as the MCP, and provides association rules that explain why certain nodes have been assigned high probabilities of being part of a large clique by the GNN. 
The graph instances and node features that we considered, a short introduction to FP-Growth, the ARM algorithm that we used, and computational considerations for the numerical experiments are now discussed. 

\subsection{Graph instances}
\begin{figure}[ht!]
    \centering
    \includegraphics[width=1.0\textwidth]{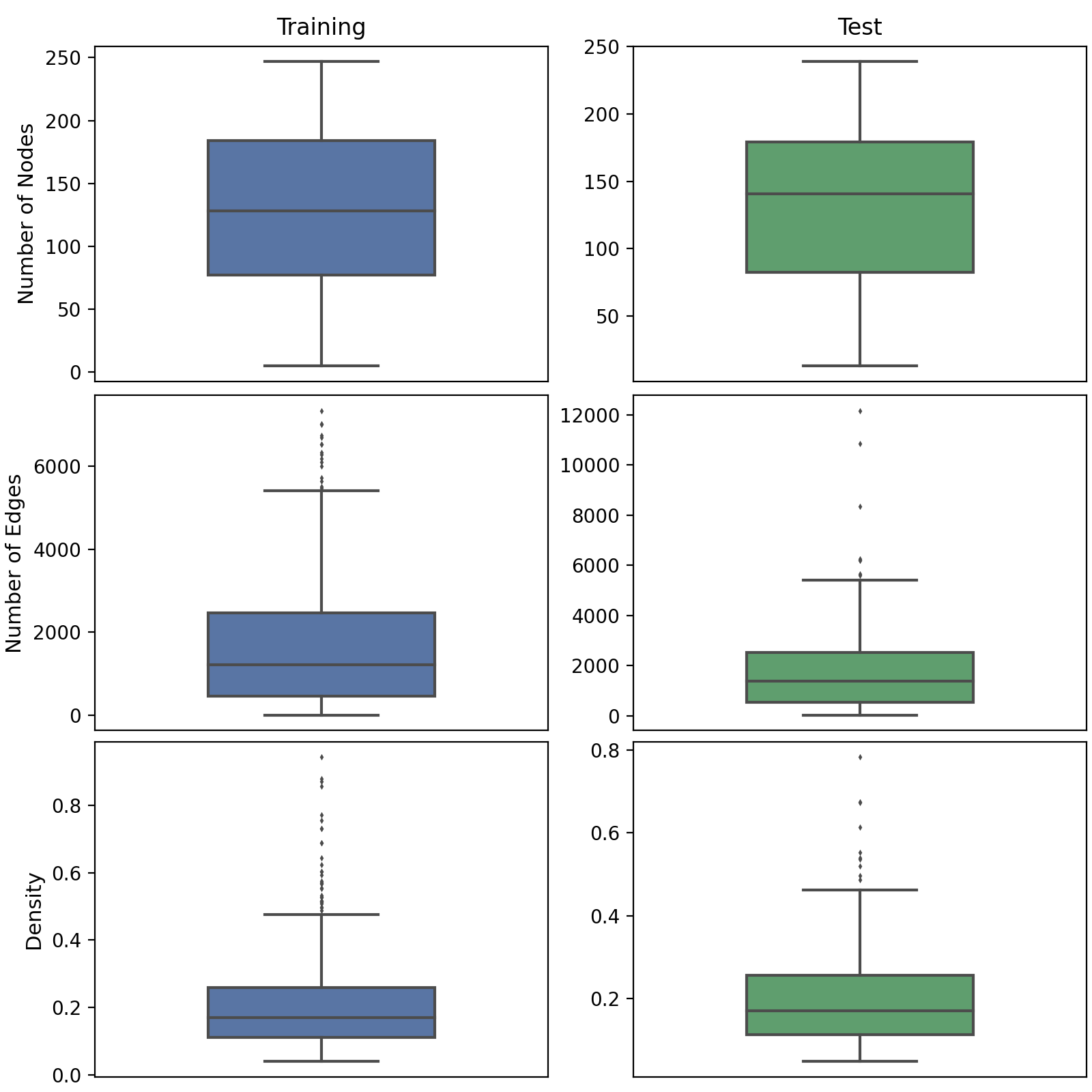}
    \caption{Node count, edge count, and density distribution of 583 training and 196 test graph instances of the TWITTER dataset}
    \label{fig:TWITTER_graphs_summary}
\end{figure}

\begin{figure}[ht!]
    \centering
    \includegraphics[width=1.0\textwidth]{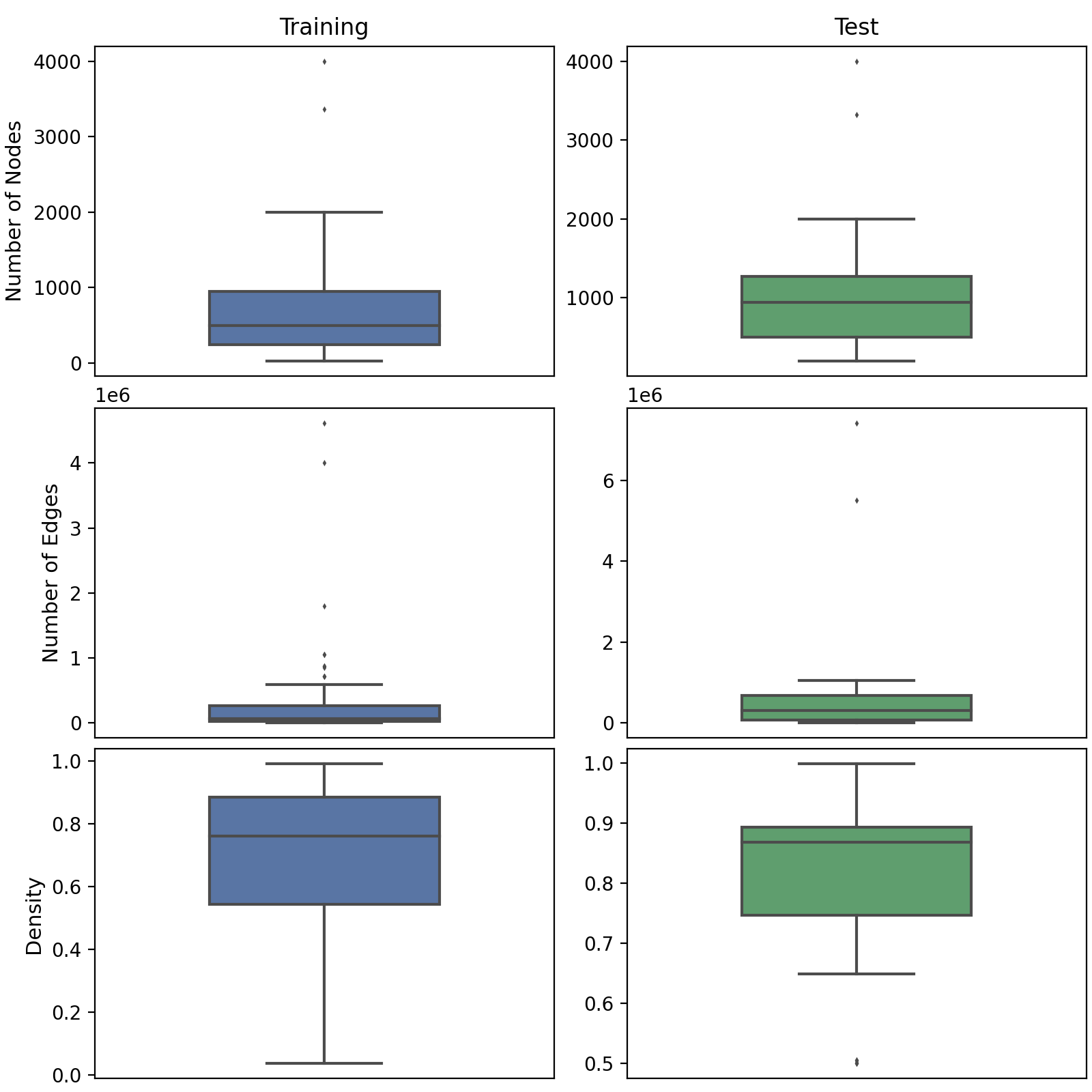}
    \caption{Node count, edge count, and density distribution of 76 training and 34 test graph instances of the BHOSLIB and DIMACS datasets}
    \label{fig:BHOSLIB_DIMACS_graphs_summary}
\end{figure}

The analysis was performed on two benchmark graph datasets: TWITTER \cite{snapnets} and BHOSLIB and DIMACS \cite{bhoslib_dataset}, which are well-known benchmarks in the GNN and combinatorial optimization research communities, respectively. A summary of these instances is provided in Table \ref{tab:summary_of_graphs_gnn_explanation}.A 60:20:20 split of the dataset yielded the training, validation, and test sets for the TWITTER graphs. For the BHOSLIB and DIMACS datasets, which had a comparatively fewer number of instances and greater variation in the size of graphs, the split was performed in a customized manner to ensure that graphs of different sizes were represented proportionately in the resulting training, validation, and test sets. The test set graphs of the BHOSLIB and DIMACS datasets are listed in table \ref{tab:largest_clique_sizes_2_9_features_bhoslib_dimacs_test_set}. While BHOSLIB and DIMACS constitute two separate benchmark datasets, they were considered together for this analysis, as these graphs are much larger in size and, more importantly, have a different topology compared to TWITTER graphs. For instance, all nodes across all the graphs of the BHOSLIB and DIMACS sets have a constant eccentricity (of 2), whereas it varies considerably (from 0 to 10, with a median value of 4) across nodes for TWITTER graphs. BHOSLIB and DIMACS graphs are also much denser compared to TWITTER graphs. Figures \ref{fig:TWITTER_graphs_summary} and \ref{fig:BHOSLIB_DIMACS_graphs_summary} show the node count, edge count, and density distributions of the training and test datasets from these two benchmarks, respectively.         

\begin{table}[h!]
    \centering
    \caption{Summary of graph instances datasets used for demonstrating ARM-Explainer on the GNN predictions for the maximum clique problem}
    \begin{tabularx}{\textwidth}{>{\centering\arraybackslash}X >{\centering\arraybackslash}X >{\centering\arraybackslash}X}
        \hline
        \textbf{Dataset Name} & \textbf{Num.Training Graphs} & \textbf{Num.Test Graphs} \\
        \hline
        TWITTER & 583 & 196 \\
        \hline
        BHOSLIB-DIMACS & 76 & 34 \\
        \hline
    \end{tabularx}
    \label{tab:summary_of_graphs_gnn_explanation}
\end{table}

\subsection{Node features}
\label{subsec: node_features}

The authors of HGS have considered three node features, namely eccentricity, the clustering coefficient, and the logarithm of the node degree, as inputs to the first layer of their GNN. Similarly, authors of GCON \cite{wenkel2024towards} have used node degree, eccentricity, clustering coefficient, and triangle counts as inputs. However, there are several other informative node features through which a GNN can potentially learn rich topological information and improve its predictions for graph-based COPs. Indeed, we observed this to be the case, particularly for the large graphs of the BHOSLIB and DIMACS datasets, where incorporating additional node features led to a significant improvement in the size of the largest clique determined by the GNN. \\          

Ten node features were considered in this study, namely the logarithm of node degree, the logarithm of the number of triangles, clustering coefficient, eccentricity, betweenness centrality, closeness centrality, degree centrality, eigenvector centrality, the logarithm of the median of the degrees of neighboring nodes, and the logarithm of the standard deviation of the degrees of neighboring nodes. The selection of an additional four centrality-based features and two neighboring-node-based features was motivated by the reasonable assumption that nodes belonging to large cliques may have values for some of these features that are significantly different from those of the rest of the nodes in the graph—an assumption that proved to hold true, as will be shown in Section \ref{sec:results} when we discuss the results. The time complexities of these features are listed in Table \ref{tab:node_features_used_in_study} of the Appendix. All these features are computable in polynomial time, and there are optimized recipes available to compute several of them much faster than the indicated worst case time-complexities (mentioned in the notes accompanying Table \ref{tab:node_features_used_in_study}. The correlations between node features for the two datasets are shown in the heatmaps of Figures \ref{fig:TWITTER_node_feature_correlation_plot} and \ref{fig:BHOSLIB_DIMACS_node_feature_correlation_plot}, respectively, in the Appendix. It can be observed from these heatmaps that there is a significant difference between the correlation values of several node feature pairs across the two datasets, again highlighting the fact that the graphs of these two datasets have different topologies.   

\subsection{FP-Growth association rule mining algorithm}
\label{sec:fpg}

ARM is an important class of techniques in data mining that uncovers interesting relationships, correlations, or co-occurrences among items within large datasets. It is widely used in domains such as market basket analysis, bioinformatics, web usage mining, and social network analysis. The objective is to identify rules of the form \( X \Rightarrow Y \), where \( X \) and \( Y \) are disjoint itemsets, indicating that the presence of \( X \) in a transaction implies a strong likelihood of \( Y \) also occurring. For example, in a retail dataset, the rule \(\{\text{bread, butter}\} \Rightarrow \{\text{milk}\}\) suggests that customers who buy bread and butter are likely to buy milk as well. Such patterns provide actionable insights for decision-making, including cross-selling, inventory management, and recommendation systems. ARM serves as one of the important classes of methods to explain observations in both commercial and scientific contexts. Several ARM algorithms, such as Apriori \cite{srikant1996mining}, ECLAT \cite{zaki1997parallel}, RELim \cite{borgelt2005keeping}, CHARM \cite{zaki2002charm}, and FP-Growth \cite{han2000mining}, among others, have been proposed in the literature. In this work, we have used FP-Growth to explain GNN predictions.\\  

The FP-Growth algorithm proposed by \cite{han2000mining} is a fast, memory-aware method for mining frequent itemsets that underpin association rules. In contrast to candidate-generation approaches such as Apriori, FP\textendash Growth compresses the database into a compact FP-tree in two passes: (i) a scan to compute item supports and prune infrequent items; (ii) a second scan to insert transactions in a fixed frequency order so that common prefixes are merged. The rule mining process proceeds by \emph{conditional pattern bases}: for each item, FP\textendash Growth builds a conditional FP-tree from transactions that contain it and recursively ``grows'' longer patterns. This divide-and-conquer process explores only relevant regions of the search space, avoiding the combinatorial explosion typical of level-wise algorithms.

Compared to Apriori, FP\textendash Growth eliminates the costly generate-and-test loop and repeated full scans, which is especially advantageous for dense data or low support thresholds. Relative to Eclat, which relies on vertical TID-list intersections, FP\textendash Growth often yields better cache behavior and lower memory requirements when many items share prefixes (in dense or moderately sparse data regimes), while still remaining competitive in sparse data. Algorithms such as Relim or H\textendash Mine share recursive ideas, but the FP-tree's structured compression and header links typically enable faster conditional exploration with fewer intermediate structures.

As an explainable machine learning tool, FP\textendash Growth pairs naturally with post-hoc explanation pipelines. Association rules derived from frequent itemsets provide transparent, human-readable statements parameterized by measures such as support, confidence, and lift that map complex model behavior to simple feature co-occurrences. In graph-based combinatorial optimization, such as the maximum clique problem, ``items'' can represent a combination of local structural indicators (e.g., percentile bins for node degrees, triangle counts, clustering coefficients, etc. ) and predictions from algorithms such as GNN, including the probability of a node belonging to a maximum clique. Applying FP\textendash Growth to these ``transactions'' surfaces recurrent substructures— for instance, rules of the form
\[
\{\text{high triangle density}\}\Rightarrow\{\text{high probability of belonging to a large clique}\}
\]
that succinctly describes the conditions under which the GNN selects nodes for large clique membership.

FP\textendash Growth's two-scan construction, absence of candidate generation, and focused conditional mining make it well-suited for large graph datasets, where exhaustive enumeration of subgraphs is infeasible. The resulting rule sets are compact yet expressive, facilitating auditable insights (what structural signals the model relies on), model debugging (identifying spurious correlations), and knowledge transfer (turning opaque embeddings into domain rules). Readers interested in learning the algorithmic details of FP-Growth are referred to \cite{han2000mining}. Thus, FP\textendash Growth provides an effective bridge between scalable pattern mining and explainable reasoning for GNN-driven solutions to graph combinatorial optimization tasks, and it motivates the empirical analysis that follows.

\subsection{ARM-Explainer Methodology}
\label{sec:ARM-Explainer_methodology}

Figure \ref{fig:ARM_Explainer_Schematic} provides a schematic of the ARM-Explainer. The process of deriving explainable association rules from the node features involves the following three steps: 
\begin{enumerate}
    \item Percentile Binning: The node features, as well as the probability of a node belonging to a maximum clique, as predicted by the GNN, are both binned into five percentile categories: 0-20,20-40,40-60,60-80, and 80-100 at the graph instance level. In other words, all the node features and the node output probabilities are transformed from their numerical values to categorical percentile values for each graph instance, with the top and bottom 20 percentile values for that graph being placed in the 80-100 and 0-20 percentile levels, respectively. While there can be other binning choices, our study found that binning by 20 percentile increments yielded results with good support and confidence measures while avoiding overfitting. It should be noted that only the nodes in the top and bottom 20 percentiles (i.e., 40 percent of the total nodes in a graph) in terms of their predicted probability values of belonging to a maximum clique from each graph instance were selected. This was done with the aim of identifying the node features that play an important role for nodes that the GNN considers to be either very likely or very unlikely to be members of a maximum clique.  
    \item Setting FP-Growth Hyperparameters: It is important to carefully select the hyperparameters of the FP-Growth algorithm to avoid generating trivial rules and a combinatorial explosion of rules. We selected a minimum support of 0.05 and a minimum confidence of 0.1 for an association rule to be selected. (Definitions of the key metrics of association rule mining are provided in \ref{sec:arm_terminology_definitions}. The values of support and confidence for several rules were well above the threshold, as will be discussed in Section \ref{sec:results}.
    \item Rule Selection: ARM can generate hundreds of rules, and several of these rules can overlap with one another. We developed an algorithm, Greedy Selection of Non-Overlapping Association Rules (see Algorithm \ref{alg:nonoverlap-rules}), to select non-overlapping association rules from the output. This algorithm arranges the rules in decreasing order of support and subsequently adds non-overlapping rules greedily, one-by-one, until all rules in the initial set are covered.
    
\end{enumerate}

\begin{algorithm}[t]
\DontPrintSemicolon
\caption{Greedy Selection of Non-Overlapping Association Rules}
\label{alg:nonoverlap-rules}
\KwIn{
\begin{tabular}{@{}l}
$\mathcal{R}$: set of rules (columns: \texttt{antecedents}, \texttt{consequents}, \\
\hspace{1.7em} and numeric metrics such as \texttt{support}, \texttt{confidence}, \texttt{lift} \\
$c^\star$: target consequent value to retain \\
$m$: name of sort metric (e.g., \texttt{support}, \texttt{confidence}, \texttt{lift} \\
$\varepsilon = 0.01$: tolerance for treating adjacent intervals as disjoint
\end{tabular}
}
\KwOut{$\mathcal{S}$: ordered list of selected non-overlapping rules}

\SetKwFunction{Parse}{ParseAntecedents}
\SetKwFunction{Disjoint}{IsDisjoint}
\SetKw{Continue}{continue}

$\mathcal{R} \leftarrow \{ r \in \mathcal{R} \mid r.\texttt{consequents} = c^\star \}$\;
\If{$\mathcal{R}=\varnothing$}{\Return $\varnothing$}

\tcp{Sort rules descending by the chosen metric}
\(\mathcal{R} \leftarrow \text{SortDesc}(\mathcal{R}, \text{key}=m)\)\;

$\mathcal{S} \leftarrow [\ ]$ \tcp*{selected rules (in order)}
$\mathcal{F} \leftarrow [\ ]$ \tcp*{per-selected rule parsed feature-interval lists}

\ForEach{$r \in \mathcal{R}$}{
    $\mathcal{I} \leftarrow$ \Parse{$r.\texttt{antecedents}$}\;
    \If{$\mathcal{I}=\varnothing$}{\Continue}
    \If{$\forall \mathcal{J} \in \mathcal{F}:~ \Disjoint(\mathcal{I}, \mathcal{J}, \varepsilon)$}{
        append $r$ to $\mathcal{S}$; append $\mathcal{I}$ to $\mathcal{F}$\;
    }
    \Else{
        \tcp*[l]{discard $r$ (its feature(s) overlap with existing feature interval(s) in $\mathcal{F}$)}
    }
}
\Return $\mathcal{S}$\;

\vspace{0.5ex}
\noindent\textit{Notes.} \textsc{ParseAntecedents} turns an antecedent string into a list of
feature-interval tuples $(f,\ell,u)$ with percent bounds as floats. \textsc{IsDisjoint} returns
\texttt{true} iff, for every feature shared by two lists, their intervals do not overlap beyond
$\varepsilon$; adjacency (e.g., $[0,19.99]$ vs.\ $[20,39.99]$) is treated as disjoint.
\end{algorithm}

\section{Computational experiments}

The computational experiments were performed on a Linux Ubuntu 22.04.3 LTS 64 bit system, utilizing an AMD Ryzen 1920 processor with 24 threads operating at a clock speed of 4 GHZ and a TU106 GeForce RTX 2070 GPU equipped with 8 GB of memory.\\

The HGS model was trained on the TWITTER and BHOSLIB-DIMACS graphs separately, and the models providing the highest average clique sizes across all instances were selected as the models to be used for explainable machine learning using the ARM-Explainer. Four
computational runs were performed—one each for the TWITTER and BHOSLIB-DIMACS datasets, and one each with 3 and 10 node features, respectively. The sets of three and ten node features were the same as described in Section \ref{subsec: node_features}. Since BHOSLIB and DIMACS graphs had the same eccentricity for all instances, this feature was not included as part of the 3 and 10 input node features; therefore, the number of node features was 2 and 9 for this dataset, respectively. The optimal hyperparameters found for the trained models, as well as the average and median number of nodes in the largest cliques identified by the trained model on the test datasets, are summarized in Table \ref{tab: optimal_hyp_largest_clique_results}. It was observed that optimal results were obtained in each of the four cases when the values of the number of epochs, penalty coefficient, and learning rate were set to 12, 0.06, and 0.0004, respectively.    

\begin{table}[htbp]
\centering
\scriptsize
\caption{Optimal hyperparameters and largest clique results for TWITTER and BHOSLIB-DIMACS test datasets}
\label{tab: optimal_hyp_largest_clique_results}
\begin{tabularx}{\textwidth}{c c c c c c c}
\toprule
\makecell{\textbf{Dataset}} &
\makecell{\textbf{Features}} &
\makecell{\textbf{Epochs}} &
\makecell{\textbf{Penalty}\\\textbf{Coefficient}} &
\makecell{\textbf{Learning}\\\textbf{Rate}} &
\makecell{\textbf{Average Nodes}\\\textbf{in Largest Clique}} &
\makecell{\textbf{Median Nodes}\\\textbf{in Largest Clique}} \\
\midrule
TWITTER & 3  & 12 & 0.06 & 0.0004 & 15.11 & 13 \\
TWITTER & 10 & 12 & 0.06 & 0.0004 & 15.68 & 13 \\
BHOSLIB-DIMACS & 2  & 12 & 0.06 & 0.0004 & 68.2 & 29.5 \\
BHOSLIB-DIMACS & 9 & 12 & 0.06 & 0.0004 & 72.4 & 36 \\
\bottomrule
\end{tabularx}
\normalsize
\end{table}

\section{Results and discussion}
\label{sec:results}

As is evident from the columns for average and median node counts of the largest cliques found in Table \ref{tab: optimal_hyp_largest_clique_results}, providing a greater number of node features as inputs does lead to improvements in GNN performance. This was especially evident for large graphs of the BHOSLIB-DIMACS dataset, where the median number of nodes in the largest clique found increased by 22\% (from 29.5 to 36) across the test dataset. The instance-wise results are presented in Table \ref{tab:largest_clique_sizes_2_9_features_bhoslib_dimacs_test_set}. A positive percentage change in the size of the largest clique found was observed in 28 out of 34 test instances, and there was 1 graph instance in which a decrease was noted. Encouragingly, for graphs such as p\_hat500-2, p\_hat1000-3, and p\_hat1500-2, which have a wider node degree spread and larger cliques than uniform graphs, the percentage change in the largest clique size found was 100\% or greater. The average size of the largest clique found increased by 3.7\% (from 15.11 to 15.68) for the test graphs of the TWITTER dataset. 

\begin{table}[h]
\centering
\scriptsize
\caption{Comparison of sizes of largest cliques found by HGS using 2 and 9 node features across BHOSLIB-DIMACS test set instances.}
\label{tab:largest_clique_sizes_2_9_features_bhoslib_dimacs_test_set}
\begin{tabularx}{\textwidth}{
    c  
    c  
    c  
    c  
    c  
    c  
    c  
}
\toprule
\makecell{\textbf{Instances}} &
\makecell{\textbf{Number}\\\textbf{of Nodes}} &
\makecell{\textbf{Number}\\\textbf{of Edges}} &
\makecell{\textbf{Density}} &
\makecell{\textbf{Nodes in Largest}\\\textbf{Found Clique} \\\textbf{(2 features)}} &
\makecell{\textbf{Nodes in Largest}\\\textbf{Found Clique} \\\textbf{(9 features)}} &
\makecell{\textbf{Percentage}\\\textbf{Change} \\\textbf{in Clique Size}} \\
\midrule
brock200\_1 & 200 & 14834 & 0.7454 & 16 & 17 & 6\% \\
sanr200\_0.9 & 200 & 17863 & 0.8976 & 29 & 31 & 7\% \\
C250.9 & 250 & 27984 & 0.8991 & 30 & 36 & 20\% \\
sanr400\_0.7 & 400 & 55869 & 0.7001 & 14 & 16 & 14\% \\
dsjc500.5 & 500 & 62624 & 0.502 & 11 & 11 & 0\% \\
p\_hat500-2 & 500 & 62946 & 0.5046 & 13 & 26 & 100\% \\
gen400\_p0.9\_65 & 400 & 71820 & 0.9 & 42 & 42 & 0\% \\
gen400\_p0.9\_75 & 400 & 71820 & 0.9 & 39 & 40 & 3\% \\
frb30-15-1 & 450 & 83198 & 0.8235 & 22 & 23 & 5\% \\
frb30-15-5 & 450 & 83231 & 0.8239 & 22 & 26 & 18\% \\
frb35-17-1 & 595 & 148859 & 0.8424 & 26 & 28 & 8\% \\
frb35-17-2 & 595 & 148868 & 0.8424 & 26 & 28 & 8\% \\
brock800\_1 & 800 & 207505 & 0.6493 & 14 & 16 & 14\% \\
keller5 & 776 & 225990 & 0.7515 & 18 & 17 & -6\% \\
frb40-19-4 & 760 & 246815 & 0.8557 & 28 & 32 & 14\% \\
frb40-19-1 & 760 & 247106 & 0.8568 & 29 & 30 & 3\% \\
dsjc1000.5 & 1000 & 249826 & 0.5002 & 11 & 12 & 9\% \\
p\_hat1000-3 & 1000 & 371746 & 0.7442 & 26 & 58 & 123\% \\
frb45-21-1 & 945 & 386854 & 0.8673 & 32 & 36 & 13\% \\
frb45-21-5 & 945 & 387461 & 0.8687 & 32 & 34 & 6\% \\
hamming10-2 & 1024 & 518656 & 0.9902 & 500 & 511 & 2\% \\
MANN\_a45 & 1035 & 533115 & 0.9963 & 293 & 293 & 0\% \\
p\_hat1500-2 & 1500 & 568960 & 0.5061 & 20 & 46 & 130\% \\
frb50-23-3 & 1150 & 579607 & 0.8773 & 37 & 38 & 3\% \\
frb50-23-4 & 1150 & 580417 & 0.8785 & 36 & 42 & 17\% \\
frb53-24-1 & 1272 & 714129 & 0.8834 & 39 & 40 & 3\% \\
frb53-24-5 & 1272 & 714130 & 0.8834 & 39 & 42 & 8\% \\
frb56-25-5 & 1400 & 869699 & 0.8881 & 39 & 44 & 13\% \\
frb56-25-2 & 1400 & 869899 & 0.8883 & 41 & 44 & 7\% \\
C2000.5 & 2000 & 999836 & 0.5002 & 12 & 13 & 8\% \\
frb59-26-2 & 1534 & 1049648 & 0.8927 & 42 & 45 & 7\% \\
frb59-26-5 & 1534 & 1049829 & 0.8929 & 45 & 45 & 0\% \\
MANN\_a81 & 3321 & 5506380 & 0.9988 & 654 & 654 & 0\% \\
frb100-40 & 4000 & 7425226 & 0.9284 & 68 & 74 & 9\% \\
\bottomrule
\end{tabularx}
\normalsize
\end{table}

A total of 60 association rules across both datasets were generated by the ARM-Explainer, and those with the highest lift values are presented in Table \ref{tab:interpretable_association_rules_table}. The confidence values for all the rules were well above the 0.1 threshold. The support values were also well above the 0.05 threshold for 7 of the 8 rules. To illustrate the explainability of these rules, let's consider the first rule in the table. Here, for the BHOSLIB-DIMACS test instances, 3 node features are provided as inputs. There are two entries in the Consequent columns, namely MC\_Prob\_Top\_20P and MC\_Prob\_Bottom\_20P, that represent the predicted maximum clique membership probability in the top and bottom 20 percentiles, respectively. The Antecedent columns contain node-feature-based entries along with their percentiles. So, the Log Degree in [0\%, 20\%] refers to the nodes that have the logarithm of their degrees within the 0 to 20 percentile range for their particular graph instances. So, as per the first rule, for BHOSLIB-DIMACS graphs, whenever the logarithm of the degrees of nodes falls in the 0 to 20 percentile bin (i.e., nodes having low degrees), there is almost a 50\% confidence that the probabilities of these nodes being part of a maximum clique lie within the top 20 percentile. A lift value of 2.497 means that it is almost 2.5 times more likely that whenever the logarithm of the degrees of a node falls within the 0 to 20 percentile bin, it will have a probability of being part of a maximum clique lying within the top 20 percent as compared to the average likelihood of all nodes that have their probabilities of being part of a maximum clique lying within either the top or bottom 20 percentiles. \\
There are two interesting insights from the explainable rules tabulated in Table \ref{tab:interpretable_association_rules_table}:
\begin{enumerate}
    \item When we compare rules across datasets, with the number of input features being 3 (or 2 for BHOSLIB-DIMACS), we observe that, in the case of TWITTER, the nodes that have high degrees (80th to 100th percentile) are much more likely to be classified by the GNN as nodes belonging to a maximum clique. The nodes that have low degrees (0th to 20th percentile) are more likely to be classified as not belonging to a maximum clique. This is in line with the custom loss function of Equation \ref{eq:custom_loss_function_for_mcp}, which favors highly connected nodes as members of a maximum clique. On the other hand, for BHOSLIB-DIMACS, we observe that the situation is reversed. Here, the nodes that have low degrees (0 to 20 percentile) are more likely to be classified as belonging to a maximum clique, and vise versa. The same conclusion also holds for TWITTER graphs with 10 features and BHOSLIB-DIMACS graphs with 9 features, where antecedents are again different across these two datasets for the same node prediction. This finding shows that ARM-Explainer is able to extract dataset specific rules that highlight the features that the GNN considers important when arriving at its predictions. 
    \item When we compare rules within the same dataset with different numbers of input features, we observe that, in the case of BHOSLIB-DIMACS, when the number of features is increased from 2 to 9, the antecedent for nodes more likely to be classified as belonging to a maximum clique changes from nodes having low degrees (0th to 20th percentile) to a high standard deviation of the degrees of the neighboring nodes (80th to 100th percentile). Such a change in the antecedent is also observed for nodes that are more likely to be classified as not belonging to a maximum clique. As is evident from the correlation heatmap in Figure \ref{fig:BHOSLIB_DIMACS_node_feature_correlation_plot}, the logarithm of the node degree and the logarithm of the standard deviation of the degrees of neighboring nodes are uncorrelated, with a correlation coefficient of 0.04. Therefore, the GNN has learned new information about the nodes through this additional feature and considers it important for the graphs in this dataset. On the other hand, for graphs of the TWITTER dataset, where the improvement in predictive performance has been marginal (3.7\% average increase in the number of nodes in the largest predicted clique), it is observed that while there is a change in the antecedent, the features in the antecedents corresponding to nodes more likely to be classified as belonging to a maximum clique, namely the logarithm of the node degree and the logarithm of the number of triangles, are highly correlated (correlation of 0.98), as evident in Figure \ref{fig:TWITTER_node_feature_correlation_plot}. A similar conclusion also holds for nodes that are more likely to be classified as not belonging to a maximum clique. Therefore, the effect of adding additional features on predictive performance depends on the graph dataset in question, and additional features seem to significantly improve performance on larger graph instances.            
    
\end{enumerate}

\begin{table}[htbp]
\centering
\scriptsize
\caption{Explainable association rules with highest lift values discovered by ARM-Explainer for the maximum clique problem for BHOSLIB-DIMACS and TWITTER test graph instances.}
\label{tab:interpretable_association_rules_table}
\begin{threeparttable}
\makebox[\linewidth][c]{%
\begin{tabularx}{1.2\textwidth}{
    c
    c
    c
    c
    c
    c
    c
}
\toprule
\makecell{\textbf{Dataset}} &
\makecell{\textbf{Features}} &
\makecell{\textbf{Antecedents}} &
\makecell{\textbf{Consequents}} &
\makecell{\textbf{Support}} &
\makecell{\textbf{Confidence}} &
\makecell{\textbf{Lift}} \\
\midrule
BHOSLIB-DIMACS & 2  & Log Degree in [0\%, 20\%] & MC\_Prob\_Top\_20P\tnote{a}    & 0.085 & 0.499 & 2.497 \\
BHOSLIB-DIMACS & 2  & Log Degree in [80\%, 100\%] & MC\_Prob\_Bottom\_20P\tnote{b} & 0.085 & 0.496 & 2.479 \\
BHOSLIB-DIMACS & 9 & Log Std Neighbor Degree in [80\%, 100\%] & MC\_Prob\_Top\_20P    & 0.085 & 0.491 & 2.454 \\
BHOSLIB-DIMACS & 9 & Log Std Neighbor Degree in [0\%, 20\%]  & MC\_Prob\_Bottom\_20P & 0.080 & 0.477 & 2.388 \\
TWITTER          & 3  & Log Degree in [80\%, 100\%] & MC\_Prob\_Top\_20P    & 0.058 & 0.283 & 1.414 \\
TWITTER          & 3  & Log Degree in [0\%, 20\%]  & MC\_Prob\_Bottom\_20P & 0.094 & 0.481 & 2.403 \\
TWITTER          & 10 & Log Number of Triangles in [80\%, 100\%] & MC\_Prob\_Top\_20P    & 0.067 & 0.327 & 1.636 \\
TWITTER          & 10 & Eigenvector Centrality in [0\%, 20\%] & MC\_Prob\_Bottom\_20P & 0.101 & 0.518 & 2.588 \\
\bottomrule
\end{tabularx}%
}
\begin{tablenotes}
\scriptsize
\item[a] MC\_Prob\_Top\_20P: Predicted maximum clique membership probability in top 20 percentile.
\item[b] MC\_Prob\_Bottom\_20P: Predicted maximum clique membership probability in bottom 20 percentile.
\end{tablenotes}
\end{threeparttable}
\normalsize
\end{table}

\section{Conclusion and further research }
\label{sec:conclusion}

This study proposes ARM-Explainer, an association rule mining-based methodology, to explain the predictions of graph neural network algorithms on graph-based combinatorial optimization problems. To the best of the authors' knowledge, this is the first study that addresses the research gap in explaining machine learning results for graph-based COPs.\\
To begin with, we investigated the effect of adding additional node-level features as inputs on the predictive performance of the hybrid scattering network, one of the best performing GNN algorithms on the maximum clique problem. We found that doing so improved the median largest found clique size by 22\% for BHOSLIB-DIMACS, the large graphs benchmark dataset. This finding is significant, as several studies on GNN-based algorithms for COPs have mentioned that the performance of these algorithms deteriorates with the increasing size of graph instances \cite{min2022can}, \cite{karalias2020erdos}, \cite{wenkel2024towards}. Therefore, the performance of such algorithms may be significantly enhanced by providing informative node features as inputs. \\
We then showed how the ARM-Explainer, an ARM-based post-hoc, model-level explainer, is able to generate explainable rules for graphs of different sizes and topologies and pin-point with high confidence the node features and their values that the GNN considers important when arriving at its predictions.\\

This work opens up multiple avenues for further research. First, we have demonstrated the performance of ARM-Explainer in explaining the predictions of the HGS algorithm for the MCP. However, the node features, percentile binning, hyperparameter settings, and rule selection components of our methodology are generic; therefore, ARM-Explainer can be applied to a wide variety of learning algorithms and graph-based COPs to investigate its performance in different problem-algorithm combinations. Second, while we selected FP-Growth as the ARM algorithm in this study and provided the rationale for doing so, there are newer versions of FP-Growth proposed in the literature \cite{wu2024research}, as well as other ARM approaches such as those proposed by \citep{sabri2025association,karabulut2025pyaerial}, that can be applied to investigate whether they yield better association rules than FP-Growth. Third, there is scope to identify the most informative node features for different graph datasets through approaches such as SHAP values. Fourth, our post-hoc level explainer still treats the GNN as a black-box, and self-interpretable approaches can be developed. However, it is not yet clear how such an architecture would look in a COP context. For example, generating a sub-graph, such as a clique, and labeling it as the explanation for the MCP solves the problem but does not explain it. Finally, the expanded set of node features that we have considered is not exhaustive, and there could be other informative features that may further improve the predictive performance. The authors anticipate investigating some of these research directions in future work and hope that the findings presented here will inspire further research on explainable machine learning for graph-based COPs.\\

\textbf{CRediT authorship contribution statement} \\
\textbf{Bharat S. Sharman}: Conceptualization, Methodology, Software, Validation,
Formal analysis, Investigation, Data curation, Writing – original
draft, Writing – review \& editing, Visualization. \textbf{Elkafi Hassini}: Conceptualization,
Methodology, Project administration, Resources, Supervision, Writing – review
\& editing.\\

\textbf{Compliance with Ethical Standards}: 
\begin{enumerate}
    \item \textbf{Funding}: The authors did not receive support from any organization for the submitted work.
    \item \textbf{Conflict of Interest}:  Bharat Sharman declares that he has no conflict of interest. Elkafi Hassini declares that has no conflict of interest.
    \item \textbf{Interests}: The authors have no relevant financial or non-financial interests to disclose.
    \item \textbf{Ethical Approval}: This article does not contain any studies with human participants or animals performed by any of the authors.
\end{enumerate}

\textbf{Data Availability}
The datasets analysed during the current study are available in the following repositories: 
\begin{enumerate}
    \item TWITTER: (http://snap.stanford.edu/data)
    \item BHOSLIB and DIMACS: (https://networkrepository.com/)  
\end{enumerate} 

\textbf{Acknowledgments}

The authors also acknowledge the authors of the HGS algorithm, who made their code openly available, which was essential for this research. This research did not receive any specific grant from funding agencies in the public, commercial, or not-for-profit sectors.\\

\textbf{Declaration of generative AI and AI-assisted technologies in the writing process}

During the preparation of this work, the author(s) used Writefull for scientific writing to paraphrase their initial draft to improve its scientific language. After using this tool, the authors reviewed and edited the content as needed and take full responsibility for the content of the published article.




\bibliography{cas_refs}

@article{mansouri2025freight,
  title={Freight Gateway Consolidation for Purolator International Using Integer Programming},
  author={Mansouri, Bahareh and Abdallah, Ramy and Tamvada, Srinivas and Hassini, Elkafi and Bellino, Brian and Khelil, Khelil},
  journal={INFORMS Journal on Applied Analytics},
  volume={55},
  number={3},
  pages={263--278},
  year={2025},
  publisher={INFORMS}
}

@article{hassini2025modeling,
  title={Modeling the impact of IoT technology on food supply chain operations},
  author={Hassini, Elkafi and Ben-Daya, Mohamed and Bahroun, Zied},
  journal={Annals of Operations Research},
  volume={348},
  number={3},
  pages={1619--1648},
  year={2025},
  publisher={Springer}
}

@article{kaya2022review,
  title={A review on the studies employing artificial bee colony algorithm to solve combinatorial optimization problems},
  author={Kaya, Ebubekir and Gorkemli, Beyza and Akay, Bahriye and Karaboga, Dervis},
  journal={Engineering Applications of Artificial Intelligence},
  volume={115},
  pages={105311},
  year={2022},
  publisher={Elsevier}
}

@article{dragotto2024critical,
  title={The critical node game},
  author={Dragotto, Gabriele and Boukhtouta, Amine and Lodi, Andrea and Taobane, Mehdi},
  journal={Journal of Combinatorial Optimization},
  volume={47},
  number={5},
  pages={74},
  year={2024},
  publisher={Springer}
}

@article{kakkad2023survey,
  title={A survey on explainability of graph neural networks},
  author={Kakkad, Jaykumar and Jannu, Jaspal and Sharma, Kartik and Aggarwal, Charu and Medya, Sourav},
  journal={arXiv preprint arXiv:2306.01958},
  year={2023}
}

@article{wenkel2024towards,
  title={Towards a general recipe for combinatorial optimization with multi-filter gnns},
  author={Wenkel, Frederik and Cant{\"u}rk, Semih and Horoi, Stefan and Perlmutter, Michael and Wolf, Guy},
  journal={arXiv preprint arXiv:2405.20543},
  year={2024}
}

@inproceedings{schank2005finding,
  title={Finding, counting and listing all triangles in large graphs, an experimental study},
  author={Schank, Thomas and Wagner, Dorothea},
  booktitle={International workshop on experimental and efficient algorithms},
  pages={606--609},
  year={2005},
  organization={Springer}
}

@inproceedings{srikant1996mining,
  title={Mining quantitative association rules in large relational tables},
  author={Srikant, Ramakrishnan and Agrawal, Rakesh},
  booktitle={Proceedings of the 1996 ACM SIGMOD international conference on Management of data},
  pages={1--12},
  year={1996}
}

@article{zaki1997parallel,
  title={Parallel algorithms for discovery of association rules},
  author={Zaki, Mohammed J and Parthasarathy, Srinivasan and Ogihara, Mitsunori and Li, Wei},
  journal={Data mining and knowledge discovery},
  volume={1},
  number={4},
  pages={343--373},
  year={1997},
  publisher={Springer}
}

@inproceedings{borgelt2005keeping,
  title={Keeping things simple: finding frequent item sets by recursive elimination},
  author={Borgelt, Christian},
  booktitle={Proceedings of the 1st international workshop on open source data mining: frequent pattern mining implementations},
  pages={66--70},
  year={2005}
}

@inproceedings{zaki2002charm,
  title={CHARM: An efficient algorithm for closed itemset mining},
  author={Zaki, Mohammed J and Hsiao, Ching-Jui},
  booktitle={Proceedings of the 2002 SIAM international conference on data mining},
  pages={457--473},
  year={2002},
  organization={SIAM}
}

@article{mizuno2024finding,
  title={Finding optimal pathways in chemical reaction networks using Ising machines},
  author={Mizuno, Yuta and Komatsuzaki, Tamiki},
  journal={Physical Review Research},
  volume={6},
  number={1},
  pages={013115},
  year={2024},
  publisher={APS}
}

@book{du2022introduction,
  title = {Introduction to Combinatorial Optimization},
  author = {Du, Dingzhu and Pardalos, Panos M. and Hu, Xiaodong and Wu, Weili and others},
  year = {2022},
  publisher = {Springer}
}

@article{wu2015review,
  title = {A review on algorithms for maximum clique problems},
  author = {Wu, Qinghua and Hao, Jin-Kao},
  journal = {European Journal of Operational Research},
  volume = {242},
  number = {3},
  pages = {693--709},
  year = {2015},
  publisher = {Elsevier}
}

@incollection{karp1972,
  author = {Karp, Richard M.},
  editor = {Miller, Raymond E. and Thatcher, James W. and Bohlinger, Jean D.},
  title = {Reducibility among Combinatorial Problems},
  booktitle = {Complexity of Computer Computations: Proceedings of a Symposium on the Complexity of Computer Computations},
  year = {1972},
  publisher = {Springer US},
  address = {Boston, MA},
  pages = {85--103}
}

@article{li2017minimization,
  title = {On minimization of the number of branches in branch-and-bound algorithms for the maximum clique problem},
  author = {Li, Chu-Min and Jiang, Hua and Manyà, Felip},
  journal = {Computers \& Operations Research},
  volume = {84},
  pages = {1--15},
  year = {2017},
  publisher = {Elsevier}
}

@inproceedings{li2010efficient,
  title = {An efficient branch-and-bound algorithm based on maxsat for the maximum clique problem},
  author = {Li, Chu-Min and Quan, Zhe},
  booktitle = {Proceedings of the AAAI Conference on Artificial Intelligence},
  volume = {24},
  pages = {128--133},
  year = {2010}
}

@inproceedings{jiang2016combining,
  title = {Combining efficient preprocessing and incremental MaxSAT reasoning for MaxClique in large graphs},
  author = {Jiang, Hua and Li, Chu-Min and Manyà, Felip},
  booktitle = {Proceedings of the Twenty-Second European Conference on Artificial Intelligence},
  pages = {939--947},
  year = {2016}
}

@article{san2019new,
  title = {A new branch-and-bound algorithm for the maximum weighted clique problem},
  author = {San Segundo, Pablo and Furini, Fabio and Artieda, Jorge},
  journal = {Computers \& Operations Research},
  volume = {110},
  pages = {18--33},
  year = {2019},
  publisher = {Elsevier}
}

@article{san2023clisat,
  title = {CliSAT: A new exact algorithm for hard maximum clique problems},
  author = {San Segundo, Pablo and Furini, Fabio and {\'A}lvarez, David and Pardalos, Panos M.},
  journal = {European Journal of Operational Research},
  volume = {307},
  number = {3},
  pages = {1008--1025},
  year = {2023},
  publisher = {Elsevier}
}

@article{wu2012multi,
  title = {Multi-neighborhood tabu search for the maximum weight clique problem},
  author = {Wu, Qinghua and Hao, Jin-Kao and Glover, Fred},
  journal = {Annals of Operations Research},
  volume = {196},
  pages = {611--634},
  year = {2012},
  publisher = {Springer}
}

@inproceedings{wang2016two,
  title = {Two efficient local search algorithms for maximum weight clique problem},
  author = {Wang, Yiyuan and Cai, Shaowei and Yin, Minghao},
  booktitle = {Proceedings of the AAAI Conference on Artificial Intelligence},
  volume = {30},
  year = {2016},
  pages = {805--811}
}

@article{wang2020sccwalk,
  title = {SCCWalk: An efficient local search algorithm and its improvements for maximum weight clique problem},
  author = {Wang, Yiyuan and Cai, Shaowei and Chen, Jiejiang and Yin, Minghao},
  journal = {Artificial Intelligence},
  volume = {280},
  pages = {103230},
  year = {2020},
  publisher = {Elsevier}
}

@article{angelini2021mismatching,
  title = {Mismatching as a tool to enhance algorithmic performances of Monte Carlo methods for the planted clique model},
  author = {Angelini, Maria Chiara and Fachin, Paolo and de Feo, Simone},
  journal = {Journal of Statistical Mechanics: Theory and Experiment},
  volume = {2021},
  number = {11},
  pages = {113406},
  year = {2021},
  publisher = {IOP Publishing}
}

@article{min2022can,
  title = {Can hybrid geometric scattering networks help solve the maximum clique problem?},
  author = {Min, Yimeng and Wenkel, Frederik and Perlmutter, Michael and Wolf, Guy},
  journal = {Advances in Neural Information Processing Systems},
  volume = {35},
  pages = {22713--22724},
  year = {2022}
}

@article{sanokowski2024variational,
  title = {Variational annealing on graphs for combinatorial optimization},
  author = {Sanokowski, Sebastian and Berghammer, Wilhelm and Hochreiter, Sepp and Lehner, Sebastian},
  journal = {Advances in Neural Information Processing Systems},
  volume = {36},
  year = {2024}
}

@article{kipf2016semi,
  title={Semi-supervised classification with graph convolutional networks},
  author={Kipf, TN},
  journal={arXiv preprint arXiv:1609.02907},
  year={2016}
}

@article{hamilton2017inductive,
  title={Inductive representation learning on large graphs},
  author={Hamilton, Will and Ying, Zhitao and Leskovec, Jure},
  journal={Advances in neural information processing systems},
  volume={30},
  year={2017}
}

@article{velivckovic2017graph,
  title={Graph attention networks},
  author={Veli{\v{c}}kovi{\'c}, Petar and Cucurull, Guillem and Casanova, Arantxa and Romero, Adriana and Lio, Pietro and Bengio, Yoshua},
  journal={arXiv preprint arXiv:1710.10903},
  year={2017}
}

@misc{gurobi,
  author = {{Gurobi Optimization, LLC}},
  title = {{Gurobi Optimizer Reference Manual}},
  year = 2023,
  url = "https://www.gurobi.com"
}

@article{karalias2020erdos,
  title={Erdos goes neural: an unsupervised learning framework for combinatorial optimization on graphs},
  author={Karalias, Nikolaos and Loukas, Andreas},
  journal={Advances in Neural Information Processing Systems},
  volume={33},
  pages={6659--6672},
  year={2020}
}

@misc{snapnets,
  author       = {Jure Leskovec and Andrej Krevl},
  title        = {{SNAP Datasets}: {Stanford} Large Network Dataset Collection},
  howpublished = {\url{http://snap.stanford.edu/data}},
  month        = jun,
  year         = 2014
}

@article{han2000mining,
  title={Mining frequent patterns without candidate generation},
  author={Han, Jiawei and Pei, Jian and Yin, Yiwen},
  journal={ACM sigmod record},
  volume={29},
  number={2},
  pages={1--12},
  year={2000},
  publisher={ACM New York, NY, USA}
}

@article{yang2022mgraphdta,
  title={MGraphDTA: deep multiscale graph neural network for explainable drug--target binding affinity prediction},
  author={Yang, Ziduo and Zhong, Weihe and Zhao, Lu and Chen, Calvin Yu-Chian},
  journal={Chemical science},
  volume={13},
  number={3},
  pages={816--833},
  year={2022},
  publisher={Royal Society of Chemistry}
}

@article{pfeifer2022gnn,
  title={GNN-SubNet: disease subnetwork detection with explainable graph neural networks},
  author={Pfeifer, Bastian and Saranti, Anna and Holzinger, Andreas},
  journal={Bioinformatics},
  volume={38},
  number={Supplement\_2},
  pages={ii120--ii126},
  year={2022},
  publisher={Oxford University Press}
}

@article{metsch2024clarus,
  title={CLARUS: An interactive explainable AI platform for manual counterfactuals in graph neural networks},
  author={Metsch, Jacqueline Michelle and Saranti, Anna and Angerschmid, Alessa and Pfeifer, Bastian and Klemt, Vanessa and Holzinger, Andreas and Hauschild, Anne-Christin},
  journal={Journal of Biomedical Informatics},
  volume={150},
  pages={104600},
  year={2024},
  publisher={Elsevier}
}

@article{mastropietro2023xgdag,
  title={XGDAG: explainable gene--disease associations via graph neural networks},
  author={Mastropietro, Andrea and De Carlo, Gianluca and Anagnostopoulos, Aris},
  journal={Bioinformatics},
  volume={39},
  number={8},
  pages={btad482},
  year={2023},
  publisher={Oxford University Press}
}

@article{tian2023predicting,
  title={Predicting molecular properties based on the interpretable graph neural network with multistep focus mechanism},
  author={Tian, Yanan and Wang, Xiaorui and Yao, Xiaojun and Liu, Huanxiang and Yang, Ying},
  journal={Briefings in bioinformatics},
  volume={24},
  number={1},
  pages={bbac534},
  year={2023},
  publisher={Oxford University Press}
}

@article{aouichaoui2023application,
  title={Application of interpretable group-embedded graph neural networks for pure compound properties},
  author={Aouichaoui, Adem RN and Fan, Fan and Abildskov, Jens and Sin, G{\"u}rkan},
  journal={Computers \& Chemical Engineering},
  volume={176},
  pages={108291},
  year={2023},
  publisher={Elsevier}
}

@article{jian2022predicting,
  title={Predicting CO2 absorption in ionic liquids with molecular descriptors and explainable graph neural networks},
  author={Jian, Yue and Wang, Yuyang and Barati Farimani, Amir},
  journal={ACS Sustainable Chemistry \& Engineering},
  volume={10},
  number={50},
  pages={16681--16691},
  year={2022},
  publisher={ACS Publications}
}

@article{kotobi2023integrating,
  title={Integrating explainability into graph neural network models for the prediction of X-ray absorption spectra},
  author={Kotobi, Amir and Singh, Kanishka and H\"{o}che, Daniel and Bari, Sadia and Mei{\ss}ner, Robert H and Bande, Annika},
  journal={Journal of the American Chemical Society},
  volume={145},
  number={41},
  pages={22584--22598},
  year={2023},
  publisher={ACS Publications}
}

@article{low2022explainable,
  title={Explainable solvation free energy prediction combining graph neural networks with chemical intuition},
  author={Low, Kaycee and Coote, Michelle L and Izgorodina, Ekaterina I},
  journal={Journal of Chemical Information and Modeling},
  volume={62},
  number={22},
  pages={5457--5470},
  year={2022},
  publisher={ACS Publications}
}

@article{xie2018crystal,
  title={Crystal graph convolutional neural networks for an accurate and interpretable prediction of material properties},
  author={Xie, Tian and Grossman, Jeffrey C},
  journal={Physical review letters},
  volume={120},
  number={14},
  pages={145301},
  year={2018},
  publisher={APS}
}

@article{yang2022learning,
  title={Learning size-adaptive molecular substructures for explainable drug--drug interaction prediction by substructure-aware graph neural network},
  author={Yang, Ziduo and Zhong, Weihe and Lv, Qiujie and Chen, Calvin Yu-Chian},
  journal={Chemical science},
  volume={13},
  number={29},
  pages={8693--8703},
  year={2022},
  publisher={Royal Society of Chemistry}
}

@inproceedings{henderson2021improving,
  title={Improving molecular graph neural network explainability with orthonormalization and induced sparsity},
  author={Henderson, Ryan and Clevert, Djork-Arn{\'e} and Montanari, Floriane},
  booktitle={International Conference on Machine Learning},
  pages={4203--4213},
  year={2021},
  organization={PMLR}
}

@article{proietti2024explainable,
  title={Explainable AI in drug discovery: self-interpretable graph neural network for molecular property prediction using concept whitening},
  author={Proietti, Michela and Ragno, Alessio and Rosa, Biagio La and Ragno, Rino and Capobianco, Roberto},
  journal={Machine Learning},
  volume={113},
  number={4},
  pages={2013--2044},
  year={2024},
  publisher={Springer}
}

@inproceedings{liu2023interpretable,
  title={Interpretable chirality-aware graph neural network for quantitative structure activity relationship modeling in drug discovery},
  author={Liu, Yunchao Lance and Wang, Yu and Vu, Oanh and Moretti, Rocco and Bodenheimer, Bobby and Meiler, Jens and Derr, Tyler},
  booktitle={Proceedings of the AAAI Conference on Artificial Intelligence},
  volume={37},
  pages={14356--14364},
  year={2023}
}

@article{li2023predicting,
  title={Predicting and Interpreting Energy Barriers of Metallic Glasses with Graph Neural Networks},
  author={Li, Haoyu and Zhang, Shichang and Tang, Longwen and Bauchy, Mathieu and Sun, Yizhou},
  journal={arXiv preprint arXiv:2401.08627},
  year={2023}
}

@article{rao2022quantitative,
  title={Quantitative evaluation of explainable graph neural networks for molecular property prediction},
  author={Rao, Jiahua and Zheng, Shuangjia and Lu, Yutong and Yang, Yuedong},
  journal={Patterns},
  volume={3},
  number={12},
  year={2022},
  publisher={Elsevier}
}

@article{aouichaoui2023combining,
  title={Combining Group-Contribution concept and graph neural networks toward interpretable molecular property models},
  author={Aouichaoui, Adem RN and Fan, Fan and Mansouri, Seyed Soheil and Abildskov, Jens and Sin, Gürkan},
  journal={Journal of Chemical Information and Modeling},
  volume={63},
  number={3},
  pages={725--744},
  year={2023},
  publisher={ACS Publications}
}

@article{barwey2023multiscale,
  title={Multiscale graph neural network autoencoders for interpretable scientific machine learning},
  author={Barwey, Shivam and Shankar, Varun and Viswanathan, Venkatasubramanian and Maulik, Romit},
  journal={Journal of Computational Physics},
  volume={495},
  pages={112537},
  year={2023},
  publisher={Elsevier}
}

@inproceedings{ganz2021explaining,
  title={Explaining graph neural networks for vulnerability discovery},
  author={Ganz, Tom and H{\"a}rterich, Martin and Warnecke, Alexander and Rieck, Konrad},
  booktitle={Proceedings of the 14th ACM Workshop on Artificial Intelligence and Security},
  pages={145--156},
  year={2021}
}

@inproceedings{he2022illuminati,
  title={Illuminati: Towards explaining graph neural networks for cybersecurity analysis},
  author={He, Haoyu and Ji, Yuede and Huang, H Howie},
  booktitle={2022 IEEE 7th European Symposium on Security and Privacy (EuroS\&P)},
  pages={74--89},
  year={2022},
  organization={IEEE}
}

@inproceedings{zhu2022interpretability,
  title={Interpretability evaluation of botnet detection model based on graph neural network},
  author={Zhu, Xiaolin and Zhang, Yong and Zhang, Zhao and Guo, Da and Li, Qi and Li, Zhao},
  booktitle={IEEE INFOCOM 2022-IEEE Conference on Computer Communications Workshops (INFOCOM WKSHPS)},
  pages={1--6},
  year={2022},
  organization={IEEE}
}

@article{lo2023xg,
  title={XG-BoT: An explainable deep graph neural network for botnet detection and forensics},
  author={Lo, Wai Weng and Kulatilleke, Gayan and Sarhan, Mohanad and Layeghy, Siamak and Portmann, Marius},
  journal={Internet of Things},
  volume={22},
  pages={100747},
  year={2023},
  publisher={Elsevier}
}

@inproceedings{warmsley2022survey,
  title={A survey of explainable graph neural networks for cyber malware analysis},
  author={Warmsley, Dana and Waagen, Alex and Xu, Jiejun and Liu, Zhining and Tong, Hanghang},
  booktitle={2022 IEEE International Conference on Big Data (Big Data)},
  pages={2932--2939},
  year={2022},
  organization={IEEE}
}

@inproceedings{cao2024coca,
  title={Coca: Improving and Explaining Graph Neural Network-Based Vulnerability Detection Systems},
  author={Cao, Sicong and Sun, Xiaobing and Wu, Xiaoxue and Lo, David and Bo, Lili and Li, Bin and Liu, Wei},
  booktitle={Proceedings of the IEEE/ACM 46th International Conference on Software Engineering},
  pages={1--13},
  year={2024}
}

@article{khalid2023dfgnn,
  title={DFGNN: An interpretable and generalized graph neural network for deepfakes detection},
  author={Khalid, Fatima and Javed, Ali and Ilyas, Hafsa and Irtaza, Aun and others},
  journal={Expert Systems with Applications},
  volume={222},
  pages={119843},
  year={2023},
  publisher={Elsevier}
}

@article{amara2022graphframex,
  title={Graphframex: Towards systematic evaluation of explainability methods for graph neural networks},
  author={Amara, Kenza and Ying, Rex and Zhang, Zitao and Han, Zhihao and Shan, Yinan and Brandes, Ulrik and Schemm, Sebastian and Zhang, Ce},
  journal={arXiv preprint arXiv:2206.09677},
  year={2022}
}

@article{dai2021graph,
  title={Graph neural networks for an accurate and interpretable prediction of the properties of polycrystalline materials},
  author={Dai, Minyi and Demirel, Mehmet F and Liang, Yingyu and Hu, Jia-Mian},
  journal={npj Computational Materials},
  volume={7},
  number={1},
  pages={103},
  year={2021},
  publisher={Nature Publishing Group UK London}
}

@article{verdone2024explainable,
  title={Explainable Spatio-Temporal Graph Neural Networks for multi-site photovoltaic energy production},
  author={Verdone, Alessio and Scardapane, Simone and Panella, Massimo},
  journal={Applied Energy},
  volume={353},
  pages={122151},
  year={2024},
  publisher={Elsevier}
}

@article{gao2022interpretable,
  title={Interpretable deep learning models for hourly solar radiation prediction based on graph neural network and attention},
  author={Gao, Yuan and Miyata, Shohei and Akashi, Yasunori},
  journal={Applied Energy},
  volume={321},
  pages={119288},
  year={2022},
  publisher={Elsevier}
}

@article{li2022online,
  title={Online multi-agent forecasting with interpretable collaborative graph neural networks},
  author={Li, Maosen and Chen, Siheng and Shen, Yanning and Liu, Genjia and Tsang, Ivor W and Zhang, Ya},
  journal={IEEE Transactions on Neural Networks and Learning Systems},
  volume={35},
  number={4},
  pages={4768--4782},
  year={2022},
  publisher={IEEE}
}

@article{zhou2022identifying,
  title={Identifying user geolocation with hierarchical graph neural networks and explainable fusion},
  author={Zhou, Fan and Wang, Tianliang and Zhong, Ting and Trajcevski, Goce},
  journal={Information Fusion},
  volume={81},
  pages={1--13},
  year={2022},
  publisher={Elsevier}
}

@inproceedings{mokhtari2022echognn,
  title={EchoGNN: explainable ejection fraction estimation with graph neural networks},
  author={Mokhtari, Masoud and Tsang, Teresa and Abolmaesumi, Purang and Liao, Renjie},
  booktitle={International Conference on Medical Image Computing and Computer-Assisted Intervention},
  pages={360--369},
  year={2022},
  organization={Springer}
}

@inproceedings{christmann2023explainable,
  title={Explainable conversational question answering over heterogeneous sources via iterative graph neural networks},
  author={Christmann, Philipp and Saha Roy, Rishiraj and Weikum, Gerhard},
  booktitle={Proceedings of the 46th International ACM SIGIR Conference on Research and Development in Information Retrieval},
  pages={643--653},
  year={2023}
}

@article{li2021braingnn,
  title={Braingnn: Interpretable brain graph neural network for fmri analysis},
  author={Li, Xiaoxiao and Zhou, Yuan and Dvornek, Nicha and Zhang, Muhan and Gao, Siyuan and Zhuang, Juntang and Scheinost, Dustin and Staib, Lawrence H and Ventola, Pamela and Duncan, James S},
  journal={Medical Image Analysis},
  volume={74},
  pages={102233},
  year={2021},
  publisher={Elsevier}
}

@article{cui2021brainnnexplainer,
  title={Brainnnexplainer: An interpretable graph neural network framework for brain network based disease analysis},
  author={Cui, Hejie and Dai, Wei and Zhu, Yanqiao and Li, Xiaoxiao and He, Lifang and Yang, Carl},
  journal={arXiv preprint arXiv:2107.05097},
  year={2021}
}

@inproceedings{cui2022interpretable,
  title={Interpretable graph neural networks for connectome-based brain disorder analysis},
  author={Cui, Hejie and Dai, Wei and Zhu, Yanqiao and Li, Xiaoxiao and He, Lifang and Yang, Carl},
  booktitle={International Conference on Medical Image Computing and Computer-Assisted Intervention},
  pages={375--385},
  year={2022},
  organization={Springer}
}

@article{zheng2024ci,
  title={Ci-gnn: A granger causality-inspired graph neural network for interpretable brain network-based psychiatric diagnosis},
  author={Zheng, Kaizhong and Yu, Shujian and Chen, Badong},
  journal={Neural Networks},
  volume={172},
  pages={106147},
  year={2024},
  publisher={Elsevier}
}

@article{harl2020explainable,
  title={Explainable predictive business process monitoring using gated graph neural networks},
  author={Harl, Maximilian and Weinzierl, Sven and Stierle, Mathias and Matzner, Martin},
  journal={Journal of Decision Systems},
  volume={29},
  number={sup1},
  pages={312--327},
  year={2020},
  publisher={Taylor \& Francis}
}

@article{lin2022efficient,
  title={Efficient and interpretable robot manipulation with graph neural networks},
  author={Lin, Yixin and Wang, Austin S and Undersander, Eric and Rai, Akshara},
  journal={IEEE Robotics and Automation Letters},
  volume={7},
  number={2},
  pages={2740--2747},
  year={2022},
  publisher={IEEE}
}

@article{tygesen2023unboxing,
  title={Unboxing the graph: Towards interpretable graph neural networks for transport prediction through neural relational inference},
  author={Tygesen, Mathias Niemann and Pereira, Francisco Camara and Rodrigues, Filipe},
  journal={Transportation research part C: emerging technologies},
  volume={146},
  pages={103946},
  year={2023},
  publisher={Elsevier}
}

@inproceedings{li2023interpretable,
  title={Interpretable sparsification of brain graphs: Better practices and effective designs for graph neural networks},
  author={Li, Gaotang and Duda, Marlena and Zhang, Xiang and Koutra, Danai and Yan, Yujun},
  booktitle={Proceedings of the 29th ACM SIGKDD Conference on Knowledge Discovery and Data Mining},
  pages={1223--1234},
  year={2023}
}

@article{lyu2022knowledge,
  title={Knowledge enhanced graph neural networks for explainable recommendation},
  author={Lyu, Ziyu and Wu, Yue and Lai, Junjie and Yang, Min and Li, Chengming and Zhou, Wei},
  journal={IEEE Transactions on Knowledge and Data Engineering},
  volume={35},
  number={5},
  pages={4954--4968},
  year={2022},
  publisher={IEEE}
}

@inproceedings{shuai2023topic,
  title={Topic-enhanced graph neural networks for extraction-based explainable recommendation},
  author={Shuai, Jie and Wu, Le and Zhang, Kun and Sun, Peijie and Hong, Richang and Wang, Meng},
  booktitle={Proceedings of the 46th International ACM SIGIR Conference on Research and Development in Information Retrieval},
  pages={1188--1197},
  year={2023}
}

@article{yu2020graph,
  title={Graph information bottleneck for subgraph recognition},
  author={Yu, Junchi and Xu, Tingyang and Rong, Yu and Bian, Yatao and Huang, Junzhou and He, Ran},
  journal={arXiv preprint arXiv:2010.05563},
  year={2020}
}

@inproceedings{yu2022improving,
  title={Improving subgraph recognition with variational graph information bottleneck},
  author={Yu, Junchi and Cao, Jie and He, Ran},
  booktitle={Proceedings of the IEEE/CVF Conference on Computer Vision and Pattern Recognition},
  pages={19396--19405},
  year={2022}
}

@inproceedings{miao2022interpretable,
  title={Interpretable and generalizable graph learning via stochastic attention mechanism},
  author={Miao, Siqi and Liu, Mia and Li, Pan},
  booktitle={International Conference on Machine Learning},
  pages={15524--15543},
  year={2022},
  organization={PMLR}
}

@article{bajaj2021robust,
  title={Robust counterfactual explanations on graph neural networks},
  author={Bajaj, Mohit and Chu, Lingyang and Xue, Zi Yu and Pei, Jian and Wang, Lanjun and Lam, Peter Cho-Ho and Zhang, Yong},
  journal={Advances in Neural Information Processing Systems},
  volume={34},
  pages={5644--5655},
  year={2021}
}

@article{baldassarre2019explainability,
  title={Explainability techniques for graph convolutional networks},
  author={Baldassarre, Federico and Azizpour, Hossein},
  journal={arXiv preprint arXiv:1905.13686},
  year={2019}
}

@article{chen2022grease,
  title={Grease: Generate factual and counterfactual explanations for gnn-based recommendations},
  author={Chen, Ziheng and Silvestri, Fabrizio and Wang, Jia and Zhang, Yongfeng and Huang, Zhenhua and Ahn, Hongshik and Tolomei, Gabriele},
  journal={arXiv preprint arXiv:2208.04222},
  year={2022}
}

@inproceedings{dai2021towards,
  title={Towards self-explainable graph neural network},
  author={Dai, Enyan and Wang, Suhang},
  booktitle={Proceedings of the 30th ACM International Conference on Information \& Knowledge Management},
  pages={302--311},
  year={2021}
}

@inproceedings{duval2021graphsvx,
  title={Graphsvx: Shapley value explanations for graph neural networks},
  author={Duval, Alexandre and Malliaros, Fragkiskos D},
  booktitle={Machine Learning and Knowledge Discovery in Databases. Research Track: European Conference, ECML PKDD 2021, Bilbao, Spain, September 13--17, 2021, Proceedings, Part II 21},
  pages={302--318},
  year={2021},
  organization={Springer}
}

@inproceedings{feng2022kergnns,
  title={Kergnns: Interpretable graph neural networks with graph kernels},
  author={Feng, Aosong and You, Chenyu and Wang, Shiqiang and Tassiulas, Leandros},
  booktitle={Proceedings of the AAAI conference on artificial intelligence},
  volume={36},
  pages={6614--6622},
  year={2022}
}

@article{feng2023degree,
  title={Degree: Decomposition based explanation for graph neural networks},
  author={Feng, Qizhang and Liu, Ninghao and Yang, Fan and Tang, Ruixiang and Du, Mengnan and Hu, Xia},
  journal={arXiv preprint arXiv:2305.12895},
  year={2023}
}

@article{funke2022zorro,
  title={Zorro: Valid, sparse, and stable explanations in graph neural networks},
  author={Funke, Thorben and Khosla, Megha and Rathee, Mandeep and Anand, Avishek},
  journal={IEEE Transactions on Knowledge and Data Engineering},
  volume={35},
  number={8},
  pages={8687--8698},
  year={2022},
  publisher={IEEE}
}

@article{huang2022graphlime,
  title={Graphlime: Local interpretable model explanations for graph neural networks},
  author={Huang, Qiang and Yamada, Makoto and Tian, Yuan and Singh, Dinesh and Chang, Yi},
  journal={IEEE Transactions on Knowledge and Data Engineering},
  volume={35},
  number={7},
  pages={6968--6972},
  year={2022},
  publisher={IEEE}
}

@article{li2023dag,
  title={Dag matters! gflownets enhanced explainer for graph neural networks},
  author={Li, Wenqian and Li, Yinchuan and Li, Zhigang and Hao, Jianye and Pang, Yan},
  journal={arXiv preprint arXiv:2303.02448},
  year={2023}
}

@inproceedings{lin2021generative,
  title={Generative causal explanations for graph neural networks},
  author={Lin, Wanyu and Lan, Hao and Li, Baochun},
  booktitle={International Conference on Machine Learning},
  pages={6666--6679},
  year={2021},
  organization={PMLR}
}

@inproceedings{lucic2022cf,
  title={Cf-gnnexplainer: Counterfactual explanations for graph neural networks},
  author={Lucic, Ana and Ter Hoeve, Maartje A and Tolomei, Gabriele and De Rijke, Maarten and Silvestri, Fabrizio},
  booktitle={International Conference on Artificial Intelligence and Statistics},
  pages={4499--4511},
  year={2022},
  organization={PMLR}
}

@article{luo2020parameterized,
  title={Parameterized explainer for graph neural network},
  author={Luo, Dongsheng and Cheng, Wei and Xu, Dongkuan and Yu, Wenchao and Zong, Bo and Chen, Haifeng and Zhang, Xiang},
  journal={Advances in neural information processing systems},
  volume={33},
  pages={19620--19631},
  year={2020}
}

@article{ma2022clear,
  title={Clear: Generative counterfactual explanations on graphs},
  author={Ma, Jing and Guo, Ruocheng and Mishra, Saumitra and Zhang, Aidong and Li, Jundong},
  journal={Advances in neural information processing systems},
  volume={35},
  pages={25895--25907},
  year={2022}
}

@inproceedings{numeroso2021meg,
  title={Meg: Generating molecular counterfactual explanations for deep graph networks},
  author={Numeroso, Danilo and Bacciu, Davide},
  booktitle={2021 International Joint Conference on Neural Networks (IJCNN)},
  pages={1--8},
  year={2021},
  organization={IEEE}
}

@inproceedings{pereira2023distill,
  title={Distill n’Explain: explaining graph neural networks using simple surrogates},
  author={Pereira, Tamara and Nascimento, Erik and Resck, Lucas E and Mesquita, Diego and Souza, Amauri},
  booktitle={International Conference on Artificial Intelligence and Statistics},
  pages={6199--6214},
  year={2023},
  organization={PMLR}
}

@inproceedings{pope2019explainability,
  title={Explainability methods for graph convolutional neural networks},
  author={Pope, Phillip E and Kolouri, Soheil and Rostami, Mohammad and Martin, Charles E and Hoffmann, Heiko},
  booktitle={Proceedings of the IEEE/CVF conference on computer vision and pattern recognition},
  pages={10772--10781},
  year={2019}
}

@article{schlichtkrull2020interpreting,
  title={Interpreting graph neural networks for NLP with differentiable edge masking},
  author={Schlichtkrull, Michael Sejr and De Cao, Nicola and Titov, Ivan},
  journal={arXiv preprint arXiv:2010.00577},
  year={2020}
}

@article{schnake2021higher,
  title={Higher-order explanations of graph neural networks via relevant walks},
  author={Schnake, Thomas and Eberle, Oliver and Lederer, Jonas and Nakajima, Shinichi and Sch{\"u}tt, Kristof T and M{\"u}ller, Klaus-Robert and Montavon, Gr{\'e}goire},
  journal={IEEE transactions on pattern analysis and machine intelligence},
  volume={44},
  number={11},
  pages={7581--7596},
  year={2021},
  publisher={IEEE}
}

@article{shan2021reinforcement,
  title={Reinforcement learning enhanced explainer for graph neural networks},
  author={Shan, Caihua and Shen, Yifei and Zhang, Yao and Li, Xiang and Li, Dongsheng},
  journal={Advances in Neural Information Processing Systems},
  volume={34},
  pages={22523--22533},
  year={2021}
}

@inproceedings{tan2022learning,
  title={Learning and evaluating graph neural network explanations based on counterfactual and factual reasoning},
  author={Tan, Juntao and Geng, Shijie and Fu, Zuohui and Ge, Yingqiang and Xu, Shuyuan and Li, Yunqi and Zhang, Yongfeng},
  booktitle={Proceedings of the ACM web conference 2022},
  pages={1018--1027},
  year={2022}
}

@article{wang2021towards,
  title={Towards multi-grained explainability for graph neural networks},
  author={Wang, Xiang and Wu, Yingxin and Zhang, An and He, Xiangnan and Chua, Tat-Seng},
  journal={Advances in Neural Information Processing Systems},
  volume={34},
  pages={18446--18458},
  year={2021}
}

@article{wang2022gnninterpreter,
  title={Gnninterpreter: A probabilistic generative model-level explanation for graph neural networks},
  author={Wang, Xiaoqi and Shen, Han-Wei},
  journal={arXiv preprint arXiv:2209.07924},
  year={2022}
}

@article{wellawatte2022model,
  title={Model agnostic generation of counterfactual explanations for molecules},
  author={Wellawatte, Geemi P and Seshadri, Aditi and White, Andrew D},
  journal={Chemical science},
  volume={13},
  number={13},
  pages={3697--3705},
  year={2022},
  publisher={Royal Society of Chemistry}
}

@article{wu2022discovering,
  title={Discovering invariant rationales for graph neural networks},
  author={Wu, Ying-Xin and Wang, Xiang and Zhang, An and He, Xiangnan and Chua, Tat-Seng},
  journal={arXiv preprint arXiv:2201.12872},
  year={2022}
}

@article{ying2019gnnexplainer,
  title={Gnnexplainer: Generating explanations for graph neural networks},
  author={Ying, Zhitao and Bourgeois, Dylan and You, Jiaxuan and Zitnik, Marinka and Leskovec, Jure},
  journal={Advances in neural information processing systems},
  volume={32},
  year={2019}
}

@inproceedings{yuan2020xgnn,
  title={Xgnn: Towards model-level explanations of graph neural networks},
  author={Yuan, Hao and Tang, Jiliang and Hu, Xia and Ji, Shuiwang},
  booktitle={Proceedings of the 26th ACM SIGKDD international conference on knowledge discovery \& data mining},
  pages={430--438},
  year={2020}
}

@inproceedings{yuan2021explainability,
  title={On explainability of graph neural networks via subgraph explorations},
  author={Yuan, Hao and Yu, Haiyang and Wang, Jie and Li, Kang and Ji, Shuiwang},
  booktitle={International conference on machine learning},
  pages={12241--12252},
  year={2021},
  organization={PMLR}
}

@article{zhang2022gstarx,
  title={Gstarx: Explaining graph neural networks with structure-aware cooperative games},
  author={Zhang, Shichang and Liu, Yozen and Shah, Neil and Sun, Yizhou},
  journal={Advances in Neural Information Processing Systems},
  volume={35},
  pages={19810--19823},
  year={2022}
}

@inproceedings{zhang2021relex,
  title={Relex: A model-agnostic relational model explainer},
  author={Zhang, Yue and Defazio, David and Ramesh, Arti},
  booktitle={Proceedings of the 2021 AAAI/ACM Conference on AI, Ethics, and Society},
  pages={1042--1049},
  year={2021}
}

@inproceedings{zhang2022protgnn,
  title={Protgnn: Towards self-explaining graph neural networks},
  author={Zhang, Zaixi and Liu, Qi and Wang, Hao and Lu, Chengqiang and Lee, Cheekong},
  booktitle={Proceedings of the AAAI Conference on Artificial Intelligence},
  volume={36},
  pages={9127--9135},
  year={2022}
}

@inproceedings{muller2024graphchef,
  title={GraphChef: Decision-Tree Recipes to Explain Graph Neural Networks},
  author={M{\"u}ller, Peter and Faber, Lukas and Martinkus, Karolis and Wattenhofer, Roger},
  booktitle={The Twelfth International Conference on Learning Representations},
  year={2024}
}

@inproceedings{ji2024stratified,
  title={Stratified GNN Explanations through Sufficient Expansion},
  author={Ji, Yuwen and Shi, Lei and Liu, Zhimeng and Wang, Ge},
  booktitle={Proceedings of the AAAI Conference on Artificial Intelligence},
  volume={38},
  pages={12839--12847},
  year={2024}
}

@inproceedings{huang2024factorized,
  title={Factorized explainer for graph neural networks},
  author={Huang, Rundong and Shirani, Farhad and Luo, Dongsheng},
  booktitle={Proceedings of the AAAI conference on artificial intelligence},
  volume={38},
  pages={12626--12634},
  year={2024}
}

@inproceedings{deng2024self,
  title={Self-Interpretable Graph Learning with Sufficient and Necessary Explanations},
  author={Deng, Jiale and Shen, Yanyan},
  booktitle={Proceedings of the AAAI Conference on Artificial Intelligence},
  volume={38},
  pages={11749--11756},
  year={2024}
}

@article{lu2024goat,
  title={GOAt: Explaining graph neural networks via graph output attribution},
  author={Lu, Shengyao and Mills, Keith G and He, Jiao and Liu, Bang and Niu, Di},
  journal={arXiv preprint arXiv:2401.14578},
  year={2024}
}

@article{lu2024eig,
  title={EiG-Search: Generating Edge-Induced Subgraphs for GNN Explanation in Linear Time},
  author={Lu, Shengyao and Liu, Bang and Mills, Keith G and He, Jiao and Niu, Di},
  journal={arXiv preprint arXiv:2405.01762},
  year={2024}
}

@inproceedings{akkas2024gnnshap,
  title={GNNShap: Scalable and Accurate GNN Explanation using Shapley Values},
  author={Akkas, Selahattin and Azad, Ariful},
  booktitle={Proceedings of the ACM on Web Conference 2024},
  pages={827--838},
  year={2024}
}

@inproceedings{chen2024generating,
  title={Generating In-Distribution Proxy Graphs for Explaining Graph Neural Networks},
  author={Chen, Zhuomin and Zhang, Jiaxing and Ni, Jingchao and Li, Xiaoting and Bian, Yuchen and Islam, Md Mezbahul and Mondal, Ananda and Wei, Hua and Luo, Dongsheng},
  year={2024},
  booktitle={Forty-first International Conference on Machine Learning}
}

@article{bui2024explaining,
  title={Explaining Graph Neural Networks via Structure-aware Interaction Index},
  author={Bui, Ngoc and Nguyen, Hieu Trung and Nguyen, Viet Anh and Ying, Rex},
  journal={arXiv preprint arXiv:2405.14352},
  year={2024}
}

@article{chen2024view,
  title={View-based explanations for graph neural networks},
  author={Chen, Tingyang and Qiu, Dazhuo and Wu, Yinghui and Khan, Arijit and Ke, Xiangyu and Gao, Yunjun},
  journal={Proceedings of the ACM on Management of Data},
  volume={2},
  number={1},
  pages={1--27},
  year={2024},
  publisher={ACM New York, NY, USA}
}

@inproceedings{chhablani2024game,
  title={Game-theoretic Counterfactual Explanation for Graph Neural Networks},
  author={Chhablani, Chirag and Jain, Sarthak and Channesh, Akshay and Kash, Ian A and Medya, Sourav},
  booktitle={Proceedings of the ACM on Web Conference 2024},
  pages={503--514},
  year={2024}
}

@article{chen2024interpretable,
  title={How Interpretable Are Interpretable Graph Neural Networks?},
  author={Chen, Yongqiang and Bian, Yatao and Han, Bo and Cheng, James},
  journal={arXiv preprint arXiv:2406.07955},
  year={2024}
}

@article{chen2024feature,
  title={Feature Attribution with Necessity and Sufficiency via Dual-stage Perturbation Test for Causal Explanation},
  author={Chen, Xuexin and Cai, Ruichu and Huang, Zhengting and Zhu, Yuxuan and Horwood, Julien and Hao, Zhifeng and Li, Zijian and Hern{\'a}ndez-Lobato, Jos{\'e} Miguel},
  journal={arXiv preprint arXiv:2402.08845},
  year={2024}
}

@inproceedings{kangunr2024,
  title={UNR-Explainer: Counterfactual Explanations for Unsupervised Node Representation Learning Models},
  author={Kang, Hyunju and Han, Geonhee and Park, Hogun},
  booktitle={The Twelfth International Conference on Learning Representations},
  year={2024}
}

@article{vermainduce2024,
  title={InduCE: Inductive Counterfactual Explanations for Graph Neural Networks},
  author={Verma, Samidha and Armgaan, Burouj and Medya, Sourav and Ranu, Sayan},
  journal={Transactions on Machine Learning Research},
  year={2024}
}

@article{qiu2024generating,
  title={Generating Robust Counterfactual Witnesses for Graph Neural Networks},
  author={Qiu, Dazhuo and Wang, Mengying and Khan, Arijit and Wu, Yinghui},
  journal={arXiv preprint arXiv:2404.19519},
  year={2024}
}

@article{wu2024research,
  title={The Research on the Improvement of FP-growth Algorithm},
  author={Wu, Zixian and Fang, Gang},
  journal={Artificial Intelligence Technology Research},
  volume={2},
  number={1},
  year={2024}
}

@article{sabri2025association,
  title={Association rule mining based approach to consider users’ preferences in the energy management of smart homes},
  author={Sabri-Laghaie, Kamyar and Momayezi, Farid and Ghaleshakhani, Negar and Maroufi, Leila},
  journal={Journal of Building Engineering},
  volume={104},
  pages={112361},
  year={2025},
  publisher={Elsevier}
}

@article{karabulut2025pyaerial,
  title={PyAerial: Scalable association rule mining from tabular data},
  author={Karabulut, Erkan and Groth, Paul and Degeler, Victoria},
  journal={SoftwareX},
  volume={31},
  pages={102341},
  year={2025},
  publisher={Elsevier}
}

@article{yang2024interpretable,
  title={Interpretable machine learning for weather and climate prediction: A review},
  author={Yang, Ruyi and Hu, Jingyu and Li, Zihao and Mu, Jianli and Yu, Tingzhao and Xia, Jiangjiang and Li, Xuhong and Dasgupta, Aritra and Xiong, Haoyi},
  journal={Atmospheric Environment},
  volume={338},
  pages={120797},
  year={2024},
  publisher={Elsevier}
}

@article{hu2025dgx,
  title={DGX: Uncovering General Behavior of Deep Graph Models with Model-level Explanation},
  author={Hu, Jinlong and Liu, Jiacheng and Dong, Shoubin and Liao, Bin and Liang, Junjie and Honavar, Vasant},
  journal={IEEE Transactions on Computational Biology and Bioinformatics},
  year={2025},
  publisher={IEEE}
}

@article{zhao2025multi,
  title={A Multi-Objective Explanation Framework for Graph Neural Networks},
  author={Zhao, Yibowen and Xu, Yonghui and Wang, Di and Zhang, Yixin and Li, Qingzhong and Cui, Lizhen},
  journal={IEEE Transactions on Knowledge and Data Engineering},
  year={2025},
  publisher={IEEE}
}

@article{saha2024graphon,
  title={Graphon-Explainer: Generating Model-Level Explanations for Graph Neural Networks using Graphons},
  author={Saha, Sayan and Bandyopadhyay, Sanghamitra},
  journal={Transactions on Machine Learning Research},
  year={2024}
}

@article{gaines2025explaining,
  title={Explaining Graph Neural Networks with mixed-integer programming},
  author={Gaines, Blake B and Zhu, Chunjiang and Bi, Jinbo},
  journal={Neurocomputing},
  pages={130214},
  year={2025},
  publisher={Elsevier}
}

@article{kosan2025gcfexplainer,
  title={GCFExplainer: Global Counterfactual Explainer for Graph Neural Networks},
  author={Kosan, Mert and Huang, Zexi and Medya, Sourav and Ranu, Sayan and Singh, Ambuj},
  journal={ACM Transactions on Intelligent Systems and Technology},
  volume={16},
  number={5},
  pages={1--23},
  year={2025},
  publisher={ACM New York, NY}
}

@inproceedings{wang2024gnnboundary,
  title={GNNBoundary: Towards explaining graph neural networks through the lens of decision boundaries},
  author={Wang, Xiaoqi and Shen, Han Wei},
  booktitle={The Twelfth International Conference on Learning Representations},
  year={2024}
}

@article{sihag2023explainable,
  title={Explainable brain age prediction using covariance neural networks},
  author={Sihag, Saurabh and Mateos, Gonzalo and McMillan, Corey and Ribeiro, Alejandro},
  journal={Advances in Neural Information Processing Systems},
  volume={36},
  pages={46958--46988},
  year={2023}
}

@inproceedings{gao2024aligning,
  title={Aligning human knowledge with visual concepts towards explainable medical image classification},
  author={Gao, Yunhe and Gu, Difei and Zhou, Mu and Metaxas, Dimitris},
  booktitle={International Conference on Medical Image Computing and Computer-Assisted Intervention},
  pages={46--56},
  year={2024},
  organization={Springer}
}

@inproceedings{chang2024path,
  title={Path-based explanation for knowledge graph completion},
  author={Chang, Heng and Ye, Jiangnan and Lopez-Avila, Alejo and Du, Jinhua and Li, Jia},
  booktitle={Proceedings of the 30th ACM SIGKDD Conference on Knowledge Discovery and Data Mining},
  pages={231--242},
  year={2024}
}

@article{armgaan2024graphtrail,
  title={GraphTrail: Translating GNN predictions into human-interpretable logical rules},
  author={Armgaan, Burouj and Dalmia, Manthan and Medya, Sourav and Ranu, Sayan},
  journal={Advances in Neural Information Processing Systems},
  volume={37},
  pages={123443--123470},
  year={2024}
}

@inproceedings{yan2008mining,
  title={Mining significant graph patterns by leap search},
  author={Yan, Xifeng and Cheng, Hong and Han, Jiawei and Yu, Philip S},
  booktitle={Proceedings of the 2008 ACM SIGMOD international conference on Management of data},
  pages={433--444},
  year={2008}
}

@article{gama2018diffusion,
  title={Diffusion scattering transforms on graphs},
  author={Gama, Fernando and Ribeiro, Alejandro and Bruna, Joan},
  journal={arXiv preprint arXiv:1806.08829},
  year={2018}
}

@article{zou2020graph,
  title={Graph convolutional neural networks via scattering},
  author={Zou, Dongmian and Lerman, Gilad},
  journal={Applied and Computational Harmonic Analysis},
  volume={49},
  number={3},
  pages={1046--1074},
  year={2020},
  publisher={Elsevier}
}

@inproceedings{gao2019geometric,
  title={Geometric scattering for graph data analysis},
  author={Gao, Feng and Wolf, Guy and Hirn, Matthew},
  booktitle={International Conference on Machine Learning},
  pages={2122--2131},
  year={2019},
  organization={PMLR}
}

@article{coifman2006diffusion,
  title={Diffusion wavelets},
  author={Coifman, Ronald R and Maggioni, Mauro},
  journal={Applied and computational harmonic analysis},
  volume={21},
  number={1},
  pages={53--94},
  year={2006},
  publisher={Elsevier}
}

@inproceedings{bhoslib_dataset,
  title = {The Network Data Repository with Interactive Graph Analytics and Visualization},
  author={Ryan A. Rossi and Nesreen K. Ahmed},
  booktitle = {AAAI},
  url={https://networkrepository.com},
  year={2015}
}

@inproceedings{yanardag2015deep,
  title={Deep graph kernels},
  author={Yanardag, Pinar and Vishwanathan, SVN},
  booktitle={Proceedings of the 21th ACM SIGKDD international conference on knowledge discovery and data mining},
  pages={1365--1374},
  year={2015}
}

@inproceedings{
azzolin2023global,
title={Global Explainability of {GNN}s via Logic Combination of Learned Concepts},
author={Steve Azzolin and Antonio Longa and Pietro Barbiero and Pietro Lio and Andrea Passerini},
booktitle={The Eleventh International Conference on Learning Representations },
year={2023},
url={https://openreview.net/forum?id=OTbRTIY4YS}
}


\appendix
\clearpage 

\section{Node features}
\subsection{Computational complexity of node features}

\begin{longtable}{p{10cm} p{7cm}}
\caption{Node-based features used for explaining GNN predictions and their computational complexity for undirected and unweighted graphs}
\label{tab:node_features_used_in_study}\\
\hline
\textbf{Feature Name} & \textbf{Computational Complexity} \\
\hline
\endfirsthead

\multicolumn{2}{c}{{\bfseries \tablename\ \thetable{} -- continued from previous page}} \\
\hline
\textbf{Feature Name} & \textbf{Computational Complexity} \\
\hline
\endhead

\hline \multicolumn{2}{r}{{Continued on next page}} \\ \hline
\endfoot

\hline
\endlastfoot

Natural logarithm of node degree \footnote{When the node degree is computed using an adjacency list.} & \(O(m)\) \\[4pt]

Natural logarithm of number of triangles \footnote{Based on the work of \cite{schank2005finding}} & 
\[
O\!\left(\sum_{(u,v)\in E} \min\!\bigl(\deg(u),\, \deg(v)\bigr)\right)
\]\\[6pt]

Clustering coefficient \footnote{The computational complexity of clustering coefficient is same as that of number of triangles.} & \[
O\!\left(\sum_{(u,v)\in E} \min\!\bigl(\deg(u),\, \deg(v)\bigr)\right)
\] \\[6pt]

Eccentricity \footnote{Eccentricity of node \(v\) equals the maximum shortest-path distance from \(v\) and one breadth first search per source node has the complexity of (O\big(n+m\big)).} & \(O\big(n\cdot (n+m)\big)\)   \\[6pt]

Betweenness centrality & \(O(n\cdot (n+m))\) \\[6pt]

Closeness centrality & \(O(n\cdot (n+m))\) \\[6pt]

Degree centrality & \(O(m)\) \\[6pt]

Eigenvector centrality (power-iteration) \footnote{Eigenvector centrality is the leading eigenvector of the adjacency matrix.
Using power iteration, each step costs one sparse matrix–vector multiplication (time $O(m)$),
and convergence in $k=O\!\bigl(\log(1/\varepsilon)/(1-|\lambda_2|/\lambda_1)\bigr)$
iterations yields a total time complexity of $O(mk)$. Convergence speed depends on spectral gap and tolerance.}
 & \(O(k\cdot (n+m))\) where \(k\) is the number of power-iteration steps until convergence.\\[6pt]

Natural logarithm of median of neighbor degrees
 & \(O(n+m)\) \\[6pt]

Natural logarithm of standard deviation of neighbor degrees & \(O(n+m)\)  \\

\end{longtable}

\bigskip
\footnotesize
\noindent\textbf{Notes:}
\begin{enumerate}
  \item In the table \(n=|V|\) (number of vertices) and \(m=|E|\) (number of edges). For sparse graphs \(m\ll n^2\) and using \(n,m\) together is customary: e.g. \(O(n\cdot m)\) is written for algorithms requiring a full single-source shortest-path from each node.
  \item For triangle-related quantities, multiple algorithms exist: node-iterator, edge-iterator, forward algorithm,, and matrix-based methods. Practical complexity is data-dependent; a common bound used for optimized algorithms is \(O(m^{3/2}\) on sparse graphs.
  \item When a per-node measure needs only local neighborhood aggregates (degree, neighbor-degree mean, neighbor-degree variance), the total work across the graph is roughly linear in \(m\) (sum of adjacency lengths). Sorting per-node neighbor lists increases costs and produces the extra logarithmic factor shown for medians if we use sorting rather than selection algorithms.
  \item Centrality measures based on all-pairs shortest paths (closeness, eccentricity, betweenness via Brandes) are typically the most expensive; approximations exist (sampling, truncated BFS) to reduce costs in large graphs.
  \item Eigenvector centrality (power iteration) costs depend on iteration count \(k\); preconditioning or using sparse linear algebra libraries can accelerate convergence in practice.
\end{enumerate}
\normalsize

\subsection{Correlation between node features}

\begin{figure}[H]
    \centering
    \includegraphics[width=\textwidth]{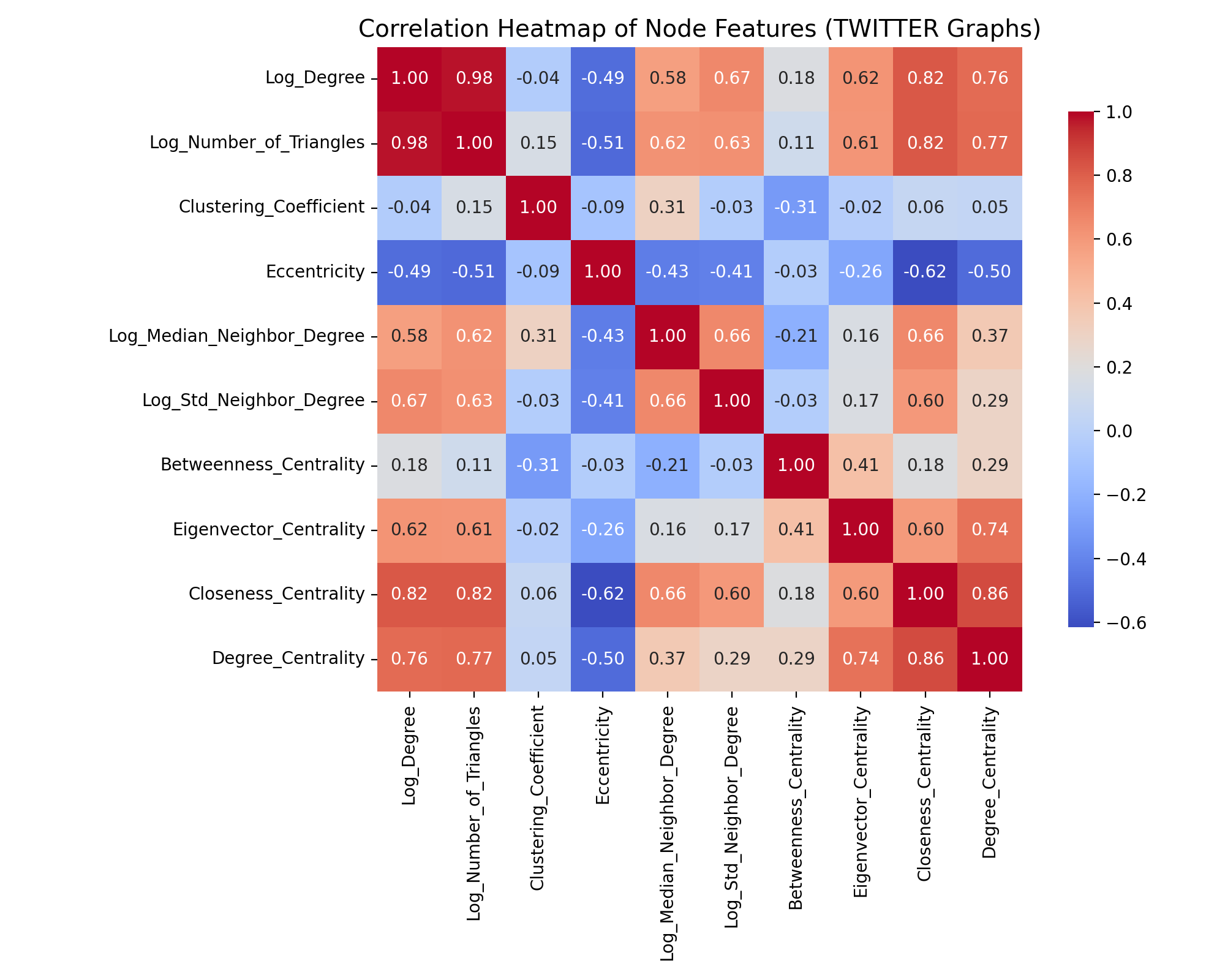}
    \caption{Correlation between node features of TWITTER graphs (train and test combined).}
    \label{fig:TWITTER_node_feature_correlation_plot}
\end{figure}

\begin{figure}[H]
    \centering
    \includegraphics[width=\textwidth]{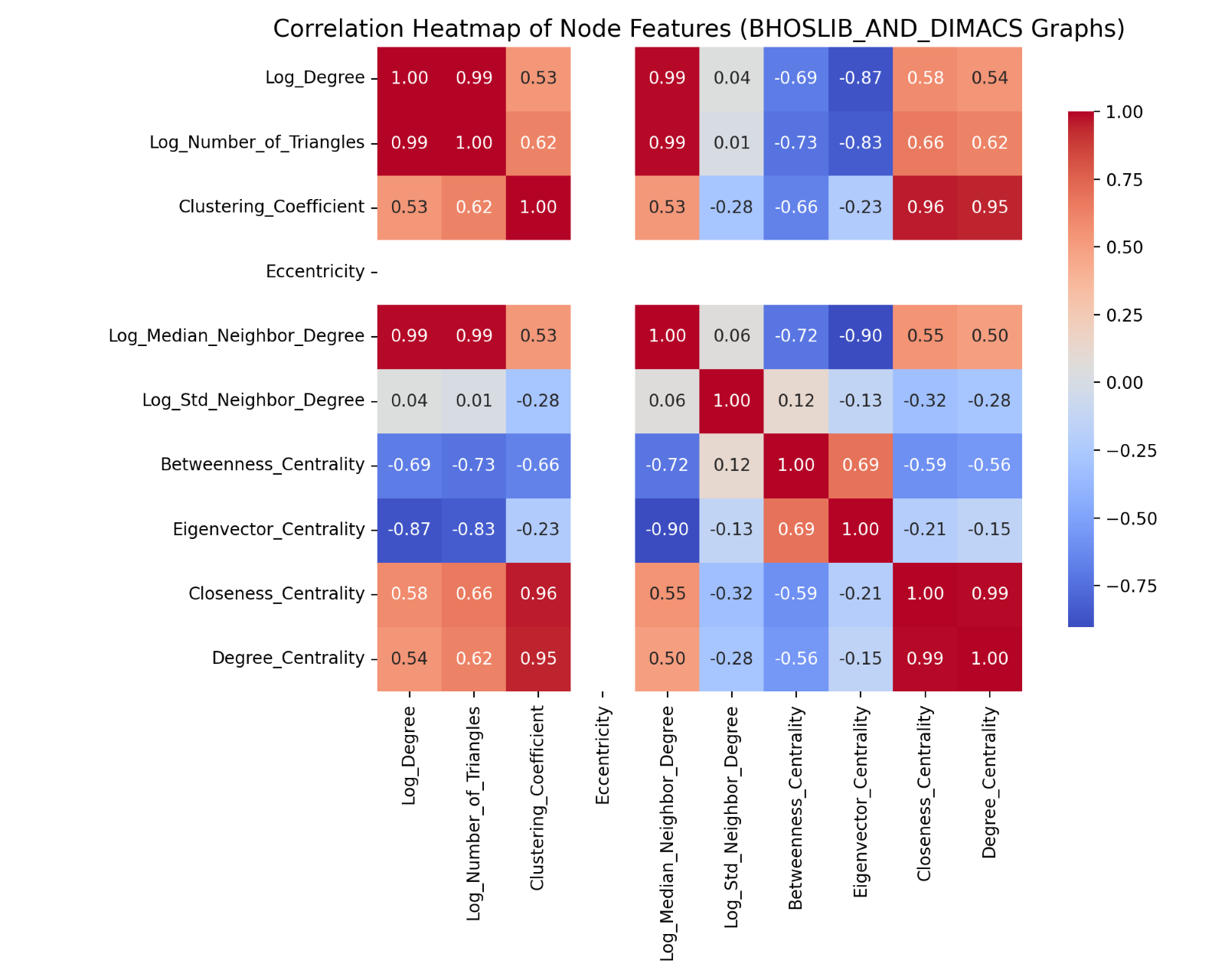}
    \caption{Correlation between node features of BHOSLIB and DIMACS graphs (train and test combined). Since eccentricity has a constant value of 2 for all graph instances in these datasets, its correlations are omitted.}
    \label{fig:BHOSLIB_DIMACS_node_feature_correlation_plot}
\end{figure}

\section{Association Rule Mining Terminology and Definitions}
\label{sec:arm_terminology_definitions}

An association rule is expressed as $X \rightarrow Y$, where $X$ (the antecedent) is a set of items and $Y$ (the consequent) is another set, indicating that the occurrence of $X$ suggests a likely occurrence of $Y$. In ARM, key metrics to assess the rule \( X \rightarrow Y \) are:

\begin{itemize}
    \item \textbf{Support}: frequency with which both \(X\) and \(Y\) occur together:
    \[
    \text{Support}(X \rightarrow Y) = \frac{\#\{\,T: X\subseteq T \land Y\subseteq T\,\}}{\#\text{transactions}}
    = P(X \cap Y).
    \]
    \item \textbf{Confidence}: how often \(Y\) appears in transactions that contain \(X\):
    \[
    \text{Confidence}(X \rightarrow Y) = \frac{P(X \cap Y)}{P(X)} = P(Y \mid X).
    \]
    \item \textbf{Lift}: how much more likely \(Y\) is to occur with \(X\) than by random chance:
    \[
    \text{Lift}(X \rightarrow Y) = \frac{P(Y \mid X)}{P(Y)} = \frac{P(X \cap Y)}{P(X)\,P(Y)}.
    \]
\end{itemize}

\end{document}